\documentclass[journal]{IEEEtran}

\usepackage{cite}
\usepackage{amsmath,amssymb,amsfonts}
\usepackage{algorithm}
\usepackage{algorithmic}
\usepackage{graphicx}
\usepackage{textcomp}
\usepackage{booktabs}
\usepackage{multirow}
\usepackage{float}
\usepackage{xcolor}
\usepackage{url}
\usepackage{bm}

\newcommand{\STAGE}[1]{\item[]\textbf{#1}}

\usepackage[hidelinks,breaklinks=true]{hyperref}

\begin{document}

\title{Null-Space Diffusion Restoration with Adaptive Uncertainty-Guided Fusion for Ultrasound Speckle Reduction}

\author{Juneyong~Lee~and~Jaeyoung~Choi,~\IEEEmembership{Member,~IEEE}%
\thanks{The authors are with the Department of Computer Science and
Engineering, Hankuk University of Foreign Studies, Yongin-si 17035,
Republic of Korea (e-mail: diziyong@hufs.ac.kr; jychoi@hufs.ac.kr).
\emph{(Corresponding author: Jaeyoung Choi.)}}%
\thanks{This work was supported by the Hankuk University of Foreign Studies
Research Fund, in part by the Development of Wave Overtopping Quantitative
Observation Technology Funded by the KIMST under Grant RS-2022-KS221567.}%
\thanks{This work has been accepted for publication in \emph{IEEE Access}.
\copyright~2026 IEEE. Personal use of this material is permitted. Permission
from IEEE must be obtained for all other uses, in any current or future media,
including reprinting/republishing this material for advertising or promotional
purposes, creating new collective works, for resale or redistribution to
servers or lists, or reuse of any copyrighted component of this work in other
works. The final version of record will be available via its DOI on IEEE
Xplore.}}

\markboth{Preprint -- accepted for publication in IEEE Access}%
{Lee \MakeLowercase{\textit{et al.}}: Uncertainty-Guided Null-Space Diffusion for Ultrasound Speckle Reduction}

\maketitle

\begin{abstract}
Ultrasound B-mode imaging commonly suffers from speckle noise and artifacts, requiring a delicate balance between contrast, resolution, and preservation of anatomical structures. Although recently developed despeckling methods have achieved some progress, supervised learning approaches remain fundamentally limited by the ground truth paradox, which arises from the absence of noise-free, ground truth reference images in \emph{in vivo} scenarios. Existing unsupervised diffusion-based methods typically enforce data consistency directly in the nonlinear log-compressed domain, which can disproportionately amplify background artifacts when mapped back to the envelope domain. To overcome these limitations, we propose an uncertainty-guided null-space diffusion (UGNS) framework, a novel label-free solution that enforces consistency correction on a stabilized positive-envelope proxy obtained via inverse log compression. The proposed UGNS introduces several technical novelties: (a) extraction of a structural prior in the stabilized envelope domain to produce a robust signal envelope that preserves anatomical structure, (b) development of an adaptive range-null reconstruction mechanism that uses an adaptive weight mask to preserve tissue regions via range-space projection, and (c) introduction of uncertainty-guided fusion in an adaptive way to mitigate sampling variability. Extensive and comparative experiments were conducted using the PICMUS benchmark and \emph{in vivo} datasets. The results demonstrate that UGNS achieves competitive generalized contrast-to-noise ratio (gCNR) values across diverse datasets. In addition, it is successfully validated that UGNS effectively suppresses speckle noise while preserving fine spatial resolution. Code is available at \url{https://github.com/yousirong/UGNS.git}.
\end{abstract}


\IEEEpeerreviewmaketitle

\section{Introduction}
\label{sec:introduction}

\IEEEPARstart{U}{ltrasound} (US) imaging has become an important medical diagnostic tool due to its noninvasive nature, real-time acquisition capabilities, and cost-effectiveness \cite{perrot2021think}. Standard B-mode US imaging is widely used for noninvasive clinical assessments, including abdominal organ evaluation, fetal monitoring, and cardiovascular diagnostics \cite{perrot2021think}. Standard B-mode US images are typically reconstructed using delay-and-sum beamforming \cite{perrot2021think}. However, the use of fixed receive focusing renders delay-and-sum (DAS) highly susceptible to speckle noise and sidelobe artifacts. Speckle noise is mainly caused by coherent interference among subresolution tissue scatterers~\cite{goodman1976some}. It inherently degrades effective spatial resolution and contrast, obscuring the boundaries of subtle anatomical structures and pathological lesions.

Much research has aimed to reduce speckle noise, evolving from classical spatial filters like Median \cite{huang1979fast} and Speckle Reducing Anisotropic Diffusion (SRAD) \cite{yu2002srad} to supervised deep learning architectures, notably ID-Net \cite{hyun2019beamforming} and Adaptive Beamforming by deep learning (ABLE) \cite{luijten2020adaptive}. Despite their efficacy, supervised paradigms consistently encounter the ground truth paradox, which arises from the fundamental absence of noise-free, ground truth reference images in \emph{in vivo} scenarios. This limitation has encouraged exploration of unsupervised, diffusion-based restoration techniques \cite{asgariandehkordi2023deep, zhang2023reconstruction, zhang2024variance}. In parallel, deep learning has demonstrated broad effectiveness across diverse engineering and medical applications \cite{jin2024lightweight,jin20253d}.

At present, most diffusion model-based approaches apply measurement guidance directly within the nonlinear log-compressed domain. Because log compression inherently amplifies low-intensity background noise, enforcing data consistency within the log-compressed domain can disrupt the approximate additivity that holds after log compression of the positive envelope, resulting in hallucinated unnatural background textures and structural artifacts.

To overcome these limitations, we propose Uncertainty-Guided Null-Space Diffusion (UGNS), a \emph{label-free} generative framework that does not require paired clean reference images or task-specific supervised fine-tuning on the target datasets. An existing diffusion-based approach \cite{kawar2022ddrm} enforces measurement consistency for general inverse problems using the SVD of an explicit linear degradation operator. However, we take a different approach by combining diffusion models with null-space consistency and uncertainty-based fusion solutions, aiming to (1) apply the diffusion prior to the normalized log-compressed domain while performing consistency correction on a stabilized positive-envelope proxy, (2) enforce range-null-space consistency through a system-matrix-free wavelet decomposition guided by an adaptive reliability mask, and (3) incorporate posterior predictive dispersion for representative-sample selection and bounded convex fusion, which enables avoiding the need to compute an explicit acoustic system matrix or to recompute its SVD when acquisition settings change.

Differing from the previous approaches \cite{zhang2023reconstruction,zhang2024variance,kawar2022ddrm}, key technical contributions of this work can be summarized in the following aspects:
\begin{itemize}
\item Development of a \textbf{label-free diffusion framework without paired clean labels} for ultrasound speckle reduction where consistency correction is applied on an inverse-compressed positive-envelope proxy rather than directly in the log-compressed domain; this enables enforcing a linear consistency projection where nonlinear domain disproportionately amplifies low-amplitude background once mapped back to the envelope.

\item Application of a system-matrix-free, spatially adaptive \emph{range-null-space reconstruction} solution to adaptively preserve tissue morphology via the use of a simplified structural prior while effectively denoising background speckle in the null space using the generative diffusion prior.

\item Introduction of a new \emph{uncertainty-guided fusion} mechanism based on posterior predictive dispersion that mitigates stochastic sampling variability while preserving fine anatomical detail.
\end{itemize}

Extensive and comparative experiments were performed using the PICMUS and \emph{in vivo} carotid benchmark datasets. A further ablation study has been performed to demonstrate the contribution of main components of our proposed UGNS, as well as to investigate the effect of hyperparameter sensitivity. Within the scope of the baselines considered in this work, UGNS achieves a favorable contrast-resolution trade-off and shows partial cross-dataset robustness under the evaluated acquisition settings. Using only a single plane-wave (1PW) acquisition, UGNS produces reconstructions that \emph{approach} the visual quality of heavily compounded DAS using 75 plane waves (75PWs) in several settings, as summarized in Table~\ref{tab:contrast_resolution} and Fig.~\ref{fig:figure2}.

\section{Related Works}
\label{sec:related_works}

\subsection{Traditional Speckle Reduction Methods}
Early ultrasound despeckling approaches, such as the Median filter \cite{huang1979fast} and Lee filter \cite{lee1980digital}, relied on spatial-domain filters that exploit local image statistics. Loizou et al. \cite{loizou2005comparative} compared conventional despeckling algorithms for carotid artery imaging and established quantitative benchmarks. While effective in reducing speckle, these filters often oversmooth fine anatomical details and edges that are important for diagnosis. Subsequently, better approaches for preserving structural information were introduced, such as SRAD, Non-Local Means, and block-matching and 3D filtering (BM3D) \cite{yu2002srad}, \cite{buades2011nonlocal}, \cite{dabov2007image}. Adaptive beamforming approaches \cite{asl2010eigenspace}, such as Eigenspace-Based Minimum Variance, further improved resolution at the image-formation stage. However, the computational cost of these methods has limited their applicability in real-world clinical settings.

\subsection{Deep Learning Methods for Ultrasound Denoising}
Convolutional neural network (CNN)-based methods have substantially advanced ultrasound denoising and beamforming. The Challenge on Ultrasound Beamforming with Deep learning \cite{bell2020cubdl} provides standardized datasets and evaluation protocols for data-driven beamformers. CNN-based approaches include ID-Net \cite{hyun2019beamforming}, which suppresses speckle in the image domain, and learning-based beamformers that predict beamforming weights, such as MobileNetV2 \cite{goudarzi2020MobileNetV2} and ABLE \cite{luijten2020adaptive}. Despite their impressive performance, these approaches require large-scale paired datasets for training, resulting in the aforementioned ground truth paradox \cite{hyun2019beamforming}, \cite{luijten2020adaptive}. As a result, their generalization to unseen clinical acquisition conditions is limited. To overcome this limitation, self-supervised methods such as Noise2Noise \cite{lehtinen2018noise2noise} and Noise2Void \cite{krull2019noise2void} have been proposed; however, these methods require task-specific paired noisy images or image-specific blind-spot pretext tasks, and tend to be sensitive to the training distribution and may overfit to synthetic speckle with hallucinated artifacts. In contrast, UGNS does not employ a self-supervised strategy of this kind: rather than constructing a task-specific pretext task from paired or single noisy images, UGNS uses an unsupervised generative diffusion prior pretrained on unlabeled ultrasound images via a standard DDPM noise-prediction objective, which is then deployed at inference time without any task-specific adaptation or paired supervision. More broadly, beyond denoising-specific approaches, recent lightweight deep architectures have shown promise in efficient medical imaging and classification tasks \cite{cakmak2025lightweight,ozel2025classification}, reflecting continued progress in deep learning for clinical applications. In addition, a few research efforts on fine-grained image enhancement have been suggested to deal with noise, blurred edges, and thermal or textural irregularities in other sensing modalities, including granulation-aided deep feature learning for far-infrared imagery \cite{paral2025adaptive} and noise-robust local-pattern descriptors for vision-sensor guidance \cite{ghosh2023histogram}.

\begin{figure*}[t!]
    \centering
    \includegraphics[width=\linewidth]{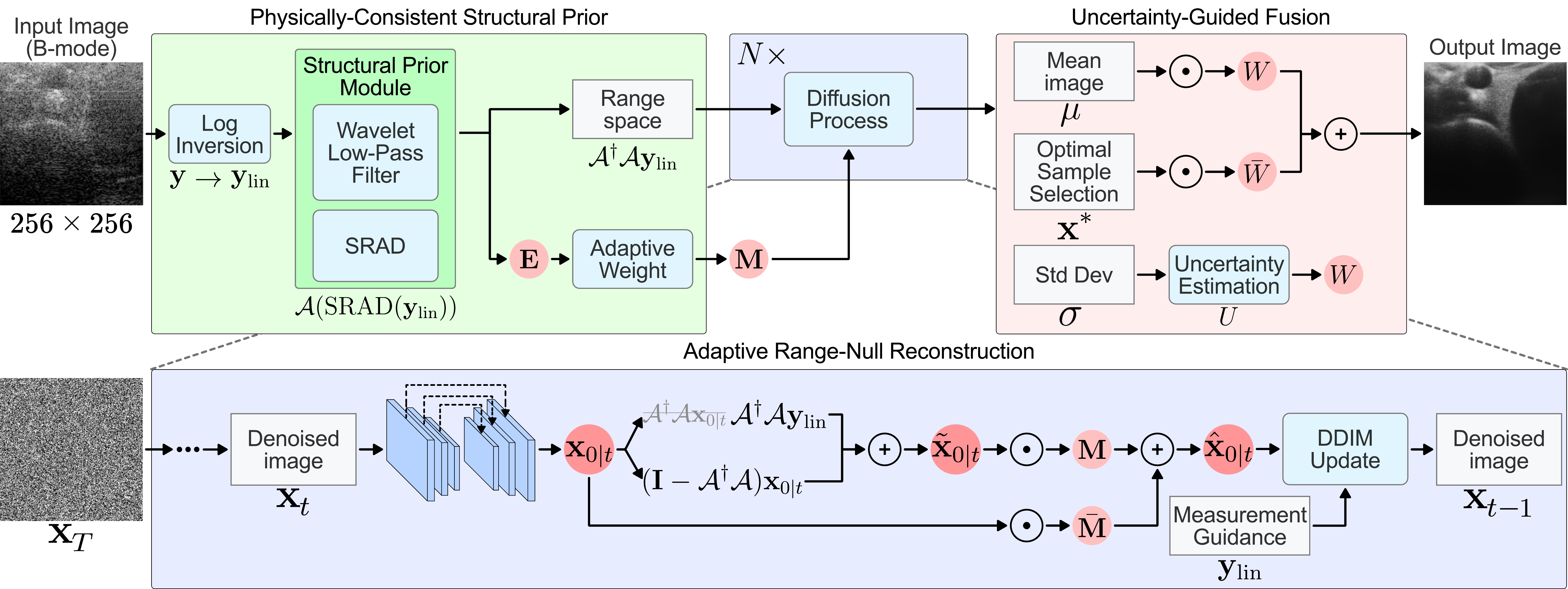}
    \caption{Overview of the proposed UGNS framework for ultrasound speckle reduction.
    The input B-mode image is first converted by Log Inversion to the stabilized positive-envelope proxy $\mathbf{y}_{\mathrm{lin}}$.
    In the Physically-Consistent Structural Prior, the Structural Prior Module combines SRAD and wavelet low-pass filtering to obtain the structural envelope $\mathbf{E}$ and the adaptive mask $\mathbf{M}$.
    Adaptive Range-Null Reconstruction then combines the range-space observation with diffusion-generated null-space detail, followed by Measurement Guidance and a DDIM Update.
    In Uncertainty-Guided Fusion, the ensemble mean $\boldsymbol{\mu}$, standard deviation $\boldsymbol{\sigma}$, uncertainty map $\mathbf{U}$, and representative sample $\mathbf{x}^*$ are fused using $\mathbf{W}$ and $\bar{\mathbf{W}}$ to produce the output image.}
    \label{fig:figure1}
\end{figure*}

\subsection{Diffusion Models for Image Restoration and Reconstruction}
Diffusion models have emerged as a powerful \emph{unsupervised generative} framework for image generation and restoration, distinct from self-supervised CNN approaches such as Noise2Noise and Noise2Void that depend on noisy image pairs or blind-spot pretext tasks. These models are highly attractive in medical imaging because they learn the data distribution from unlabeled images via a noise-prediction training objective and eliminate the need for large-scale paired training data at the task-specific fine-tuning stage. Recent ultrasound-oriented denoising diffusion probabilistic model variants \cite{asgariandehkordi2023deep}, \cite{guha2023sddpm}, \cite{li2025speckle2self}, such as DENO, Speckle Denoising Diffusion Probabilistic Models, and Speckle2Self, demonstrate improved denoising performance over CNN-based approaches. However, most diffusion-based methods apply their consistency step directly in the nonlinear log-compressed domain, which can disrupt the approximate additivity that would otherwise hold after log compression of the positive envelope. In contrast, UGNS performs consistency correction on a stabilized positive-envelope proxy obtained by inverse log compression, which helps avoid unnatural amplification of background artifacts.

General paired-data-free diffusion techniques often struggle with ultrasound-specific noise characteristics. To address general inverse problems, Denoising Diffusion Restoration Models \cite{kawar2022ddrm} introduced a singular value decomposition (SVD)-based formulation, which was extended by the Denoising Diffusion Null-Space Model (DDNM) \cite{wang2023ddnm}. DDNM preserves observations in the range space while using the diffusion prior to reconstruct missing information in the null space.

While both DDRM and DDNM have robust theoretical foundations, they are primarily designed for natural images with additive Gaussian noise, and hence lack domain-specific adaptations for ultrasound. Unlike DDRM and DDNM, \emph{UGNS actively suppresses multiplicative ultrasound speckle} by integrating a principled structural prior into the range space and mitigating stochastic variability through an uncertainty-guided fusion mechanism.

Paired-data-free diffusion techniques have been increasingly explored for ultrasound reconstruction. For example, DRUS and WDRUS \cite{zhang2023reconstruction} extend DDRM to reconstruct high-resolution images directly from radio-frequency (RF) channel data, whereas DRUSvar \cite{zhang2024variance} leverages sampling variability to derive variance maps linked to tissue reflectivity. However, \emph{the DRUS family is heavily reliant on the SVD of the acoustic system matrix.} This physical modeling is computationally expensive and requires accurate hardware parameters, such as transducer geometry and transmit waveforms. In practice, speed-of-sound inhomogeneities and device-dependent effects inevitably result in model mismatches that can degrade reconstruction performance.

Our proposed UGNS differs from other state-of-the-art diffusion model approaches \cite{asgariandehkordi2023deep,zhang2023reconstruction,zhang2024variance,kawar2022ddrm,guha2023sddpm,li2025speckle2self,wang2023ddnm} in the following ways:
\begin{itemize}
\item UGNS adopts a signal-driven, \emph{system-matrix-free} inference formulation that does not require an explicit device-specific acoustic system matrix or an SVD of such a matrix to enforce consistency.

\item UGNS utilizes the discrete wavelet transform on the already-beamformed signal to separate stochastic speckle (null space) from deterministic tissue structure (range space) based on spatial frequency, which avoids recomputing an SVD when the acquisition parameters change.

\item Because the degradation process in UGNS is defined via a frequency-domain decomposition rather than an explicit acoustic system matrix, the inference pipeline does not require computing or storing a device-specific system matrix or its SVD. This is methodologically distinct from system-matrix-dependent approaches.
\end{itemize}

\section{Proposed Method}
\label{sec:method}
Fig.~\ref{fig:figure1} presents the overall framework of the proposed UGNS method. As shown, the UGNS pipeline consists of three main components: (1) \textbf{Development of a physically-consistent structural prior} in the linear space in which by routing the signal through a structural prior module that integrates SRAD and a wavelet low-pass filter, we extract a highly robust structural envelope. This effectively suppresses speckle while strictly preserving numerical consistency. (2) \textbf{Adaptive range-null reconstruction} where by using an adaptive weight mask derived from the structural prior, high-confidence tissue regions are constrained to the low-frequency (range-space) spectrum of the observation. (3) \textbf{Uncertainty-guided fusion} is introduced to systematically mitigate sample-to-sample variability inherent in stochastic generative models. By performing comprehensive uncertainty estimation across an ensemble of diffusion samples, we execute an optimal sample selection process. The following subsections describe these components in more detail.

\subsection{Problem Formulation}
Following classical statistical acoustics for fully developed speckle~\cite{goodman1976some,burckhardt1978speckle,wagner1983statistics} and homomorphic signal processing~\cite{oppenheim1968nonlinear,achim2001novel}, we adopt a principled multiplicative envelope model~\cite{jain1989fundamentals,loupas1989adaptive}.
Let $\mathbf{i}\in\mathbb{R}_{+}^{H\times W}$ denote the observed
positive envelope, $\mathbf{s}\in\mathbb{R}_{+}^{H\times W}$ the
latent clean structural envelope, and $\mathbf{u}\in
\mathbb{R}_{+}^{H\times W}$ the multiplicative speckle field
(Rayleigh-distributed under fully developed
speckle~\cite{burckhardt1978speckle,wagner1983statistics}):
\begin{equation}
  \label{eq:mult_model}
  \mathbf{i} = \mathbf{s} \odot \mathbf{u}.
\end{equation}

Let $g(\cdot)$ denote the normalized log-compression operator
with dynamic range $D=60\,$dB, defined element-wise as:
\begin{equation}
  \label{eq:g_def}
  g(v) = 1 + \frac{20}{D}\log_{10}(v + \varepsilon),
\end{equation}
where $\varepsilon>0$ is a small stabilization constant and
the output is clipped to $[0,1]$.
Its inverse is $g^{-1}(u)=10^{D(u-1)/20}$.
We define $\mathbf{y}=g(\mathbf{i})\in[0,1]^{H\times W}$
and $\mathbf{x}=g(\mathbf{s})\in[0,1]^{H\times W}$.
Using homomorphic transformation, the multiplicative model
\eqref{eq:mult_model} becomes the exact additive identity
\emph{in the log-compressed domain}:
\begin{equation}
  \label{eq:add_model}
  \mathbf{y} = \mathbf{x} + \mathbf{n} +
               \boldsymbol{\eta}_{\mathrm{sys}},
\end{equation}
where $\mathbf{n} = g(\mathbf{u}) - g(\mathbf{1})$ is the
exact log-speckle noise term (an algebraic consequence of
applying $g$ to the multiplicative model with no approximation),
and $\boldsymbol{\eta}_{\mathrm{sys}} =
\tfrac{20}{D}\log_{10}(u_{\mathrm{sys}})$ absorbs bounded system
distortions (attenuation, aberration, and sound-speed
inhomogeneity) through the composite factor
$u_{\mathrm{sys}} = a(z)\,\alpha_{\mathrm{ab}}\,\alpha_c\,
\alpha_{\mathrm{bf}} > 0$.
This additive relation holds only in the log-compressed domain and is consistent with standard homomorphic filtering. It is not assumed to hold in the raw envelope domain. 
The inverse-compressed quantities $g^{-1}(\mathbf{y})$ and 
$g^{-1}(\mathbf{x}_{0|t})$ are interpreted as \emph{stabilized 
positive-envelope proxies} for envelope-domain consistency correction, not as estimates of the absolute acoustic pressure.
A formal derivation of the log-domain linearization, its stability 
under acoustic distortion, and four nested upper bounds on the 
residual between the UGNS signal model and the true ultrasound 
envelope are provided in the Appendix.

We further operate under the physical premise that tissue structure is predominantly governed by low-frequency components, which are strictly preserved by a low-pass operator $\mathcal{A}$ (the range space), while speckle primarily manifests in the complementary high-frequency spectrum (the null space): $\mathbf{x} \approx \mathcal{A}^\dagger \mathcal{A}\mathbf{x}$ and $\mathcal{A}\mathbf{n} \approx 0$. Numerical stability is ensured by clipping $\mathbf{y}$ and $\mathbf{x}$ to $[0,1]$, by using $\varepsilon > 0$ inside $g(\cdot)$ and $g^{-1}(\cdot)$, and by fixing $D=60$\,dB.

\subsection{Diffusion Model Preliminaries}
The diffusion prior operates in the normalized log-compressed B-mode domain $\mathbf{x} = g(\mathbf{s}) \in [0,1]^{H\times W}$, i.e., the same domain used for both pretraining and sampling. At any reverse step $t$, the clean state estimate $\mathbf{x}_{0|t}$ is predicted from the latent variable $\mathbf{x}_t$ using the optimal noise predictor $\boldsymbol{\epsilon}_{\theta}(\mathbf{x}_t, t)$:
\begin{equation}
\label{eq:x0_pred}
\mathbf{x}_{0|t} = \frac{\mathbf{x}_t - \sqrt{1-\bar{\alpha}_t}\,\boldsymbol{\epsilon}_{\theta}(\mathbf{x}_t, t)}{\sqrt{\bar{\alpha}_t}}.
\end{equation}

Although $\mathbf{x}_{0|t}$ reflects the globally learned generative prior, it lacks fidelity to the specific clinical observation. UGNS rectifies this using an intensity-adaptive consistency mechanism that is applied not directly in the log-compressed domain, but on a stabilized positive-envelope proxy obtained by inverse log compression of $\mathbf{y}$, as described below.

\subsection{UGNS Algorithm}

\subsubsection{Physically-Consistent Structural Prior}
We apply inverse log compression to obtain a \emph{stabilized positive-envelope proxy} of the observation:
\begin{equation}
  \label{eq:g_inv}
  \mathbf{y}_{\mathrm{lin}} = g^{-1}(\mathbf{y}) = 10^{D(\mathbf{y}-1)/20},
  \quad D = 60\,\mathrm{dB}.
\end{equation}
This proxy provides a stable positive-valued representation, 
bounded within the same normalized $[0,1]$ scale as 
$\mathbf{x}_{0|t}$, in which the consistency operator can be applied 
without the nonlinear amplification that would occur directly in 
the log-compressed domain.

Subsequently, we synthesize a robust structural envelope ($\mathbf{E}$) by routing the proxy through a structural prior module, which applies SRAD followed by a wavelet low-pass filter operator ($\mathcal{A}$):
\begin{equation}
\mathbf{E} = \mathcal{A}(\mathrm{SRAD}(\mathbf{y}_{\mathrm{lin}})).
\end{equation}

Here, $\mathcal{A}$ selectively extracts the LL (low-low) subband of the discrete wavelet decomposition, retaining coarse anatomical structures while attenuating high-frequency speckle components. From $\mathbf{E}$, we compute an adaptive weight mask $\mathbf{M}$ that models the local signal reliability:
\begin{equation}
\mathbf{N} = \mathrm{clip}\left( \frac{\mathbf{E} - c}{1 - c}, 0, 1 \right), \quad
\mathbf{M} = \mathbf{N}^\gamma \odot G_{\mathrm{otsu}}(\mathbf{N}),
\end{equation}

where $c$ is the center of the reliability transition and $\gamma$
controls the steepness of the mask response. Note that the values of $c$
and $\gamma$ were determined through a heuristic search based on
restoration performance. A good compromise has been found by setting $c$
in the range of $[0.35, 0.45]$, as well as by setting $\gamma$ in the
range of $[3.0, 4.0]$. Our experimental results show that varying each
parameter within this range does not alter the performance of the
proposed UGNS; hence, we set $c=0.4$, $\gamma=3.5$. $G_{\mathrm{otsu}}(\cdot)$ suppresses low-intensity background regions to reduce hallucination artifacts.

\subsubsection{Adaptive Range-Null Reconstruction}
During each iterative reverse diffusion step, measurement 
consistency is enforced by a range-null orthogonal projection 
followed by intensity-adaptive blending.

\paragraph{Range-Null Space Reconstruction}
The low-frequency (range-space) component of the stabilized
linear observation ($\mathbf{y}_{\mathrm{lin}}$) is combined
with the high-frequency (null-space) component predicted by
the diffusion model ($\mathbf{x}_{0|t}$):
\begin{equation}
\label{eq:range_null_recon}
\tilde{\mathbf{x}}_{0|t} = \mathcal{A}^\dagger \mathcal{A} \mathbf{y}_{\mathrm{lin}} + (\mathbf{I} - \mathcal{A}^\dagger \mathcal{A}) \mathbf{x}_{0|t}.
\end{equation}
Note that the range-space term in \eqref{eq:range_null_recon} is 
drawn from the linear-domain proxy $\mathbf{y}_{\mathrm{lin}}$, 
while the null-space term is obtained from the log-compressed estimate 
$\mathbf{x}_{0|t}$; this scale-aligned range-null composition allows UGNS to anchor low-frequency tissue content to envelope-domain fidelity while delegating high-frequency reconstruction to 
the diffusion prior.

\paragraph{Adaptive Mixing}
Leveraging the adaptive weight mask $\mathbf{M}$, we bias highly confident tissue regions toward the range-anchored estimate $\tilde{\mathbf{x}}_{0|t}$, while steering background regions toward the denoised prior prediction $\mathbf{x}_{0|t}$:
\begin{equation}
\hat{\mathbf{x}}_{0|t} = \mathbf{M} \odot \tilde{\mathbf{x}}_{0|t} + \bar{\mathbf{M}} \odot \mathbf{x}_{0|t},
\end{equation}
where $\bar{\mathbf{M}} = \mathbf{1} - \mathbf{M}$.

\paragraph{Measurement Guidance \& DDIM Update}
To reinforce fidelity to the acquired data, explicit measurement guidance, inspired by iterative latent variable
refinement \cite{choi2021ilvr}, is implemented using the linear
observation $\mathbf{y}_{\mathrm{lin}}$. We utilize a time-dependent coupling coefficient $\lambda_t$ modulated by a cosine schedule:
\begin{equation}
\label{eq:cosine}
\lambda_t = \lambda_{\mathrm{end}} + (\lambda_{\mathrm{start}} - \lambda_{\mathrm{end}}) \cdot \frac{1 + \cos(\pi(1 - t/T))}{2}.
\end{equation}
We formulate the guided estimate as $\hat{\mathbf{x}}_{0|t}^{\mathrm{guided}} = (1 - \lambda_t)\hat{\mathbf{x}}_{0|t} + \lambda_t \mathbf{y}_{\mathrm{lin}}$, and execute the DDIM update to compute $\mathbf{x}_{t-1}$ \cite{song2020ddim}:
\begin{equation}
\mathbf{x}_{t-1} = \sqrt{\bar{\alpha}_{t-1}} \hat{\mathbf{x}}_{0|t}^{\mathrm{guided}} + \sqrt{1 - \bar{\alpha}_{t-1}} \boldsymbol{\epsilon}_\theta(\mathbf{x}_t, t).
\end{equation}
Deterministic DDIM sampling ($\eta=0$) is configured to guarantee strict reproducibility and temporal stability in the reconstructions.

\subsubsection{Uncertainty-Guided Fusion}
To regulate generative sampling variability, we generate an ensemble of $N=10$ restored samples $\{\hat{\mathbf{x}}_0^{(n)}\}_{n=1}^N$, where each $\hat{\mathbf{x}}_0^{(n)}\in\mathbb{R}^{H\times W}$ is produced by the adaptive range-null reverse-diffusion sampler of Section~III-C2 with independent initial noise. Let $x_i^{(n)}$ denote the intensity at spatial index $i\in\{1,\dots,HW\}$ of the $n$th sample. Rather than treating uncertainty as a deterministic measurement-noise level, we define it as the \emph{posterior predictive dispersion} across the ensemble. The pixel-wise empirical posterior is:
\begin{equation}
\label{eq:emp_posterior}
\hat{p}_i(x\mid\mathbf{y}) = \frac{1}{N}\sum_{n=1}^{N}\delta\!\left(x-x_i^{(n)}\right),
\end{equation}
and its posterior mean $\mu_i$ and predictive variance $\sigma_i^2$ are:
\begin{equation}
\label{eq:mu_sigma}
\mu_i = \frac{1}{N}\sum_{n=1}^{N}x_i^{(n)}, \qquad
\sigma_i^2 = \frac{1}{N}\sum_{n=1}^{N}\left(x_i^{(n)}-\mu_i\right)^2.
\end{equation}
Here, $\sigma_i^2$ uses the biased ($N$-denominator) form for
computational convenience; the conservative upper bound in
Appendix~E is derived with $N_{\mathrm{eff}}=N-1$
(unbiased correction), which yields a slightly larger but
rigorously valid bound.
These quantities are not chosen heuristically. For any scalar estimate $a$, the bias--variance decomposition gives:
\begin{equation}
\label{eq:bias_var}
\frac{1}{N}\sum_{n=1}^{N}\left(x_i^{(n)}-a\right)^2
=
(a-\mu_i)^2+\sigma_i^2,
\end{equation}
which is minimized at $a=\mu_i$. Thus $\mu_i$ is the empirical minimum-mean-squared-error estimator under the ensemble posterior, and $\sigma_i$ directly measures the posterior spread around it. As a result, the standard deviation is a natural uncertainty proxy: it quantifies posterior dispersion in the same intensity units as the reconstruction, it does not require a separate parametric noise model in the raw envelope domain, and it is directly computable from the diffusion ensemble. The same quantity also connects to information theory: if the local marginal posterior is approximated as Gaussian, $X_i\mid\mathbf{y} \approx \mathcal{N}(\mu_i,\sigma_i^2)$, its differential entropy
\begin{equation}
\label{eq:entropy}
h(X_i\mid\mathbf{y})=\tfrac{1}{2}\log\!\left(2\pi\mathrm{e}\,\sigma_i^2\right)
\end{equation}
is monotone in $\sigma_i^2$. Thus, larger $\sigma_i$ corresponds to larger local posterior uncertainty in both the Bayesian predictive and information-theoretic senses. The normalized uncertainty map is then defined pixel-wise as:
\begin{equation}
\label{eq:U_norm}
U_i = \frac{\sigma_i}{\max_j \sigma_j+\varepsilon}, \qquad U_i \in [0,1].
\end{equation}

To select a representative structural sample from the ensemble, we score each
restored sample $\mathbf{z}^{(n)}=\hat{\mathbf{x}}_0^{(n)}$ by its
uncertainty-weighted structural sharpness as follows:
\begin{equation}
\label{eq:quality_score}
Q_n=\sum_i (1-U_i)\left\|\nabla z_i^{(n)}\right\|_2,
\end{equation}
where $\nabla$ denotes the spatial gradient operator,
$\left\|\nabla z_i^{(n)}\right\|_2$ is the local gradient magnitude at pixel
$i$, and $U_i$ is the normalized uncertainty defined in \eqref{eq:U_norm}.
Herein, $Q_n$ rewards structurally sharp samples in low-uncertainty regions
while suppressing unstable regions with large ensemble dispersion.
The representative sample is then obtained by
\begin{equation}
\label{eq:quality_selection}
n^*=\arg\max_n Q_n,\qquad \mathbf{x}^*=\mathbf{z}^{(n^*)}.
\end{equation}
This score favors structurally sharp samples in low-uncertainty regions while
down-weighting unstable regions with large ensemble dispersion.

Finally, we fuse $\mathbf{x}^*$ with the ensemble mean $\boldsymbol{\mu}$ using pixel-wise adaptive weights:
\begin{equation}
\label{eq:weight}
W_i=\mathrm{clip}\!\left(w_{\min}+(w_{\max}-w_{\min})U_i,\; w_{\min},\; w_{\max}\right),
\end{equation}
with $w_{\min}=0.7$ and $w_{\max}=0.98$, and 
$\bar{W}_i = 1 - W_i$. The final fusion is:
\begin{equation}
\label{eq:fusion}
x_{\mathrm{final},i}=W_i\,\mu_i+(1-W_i)\,x_i^*.
\end{equation}
When $U_i$ is large, $W_i$ moves toward $0.98$ and the posterior mean $\mu_i$ dominates; this is desirable because the mean is the most stable estimator under squared loss and suppresses sample-to-sample randomness. When $U_i$ is small, $W_i$ moves toward $0.7$ and the representative sharp sample retains a $30\%$ contribution; this helps preserve the edge contrast and fine anatomical detail that may be attenuated in the ensemble average. At the same time, $W_i$ never falls below $0.7$, so the fusion remains conservative.

The fusion is applied exactly once after sampling as a single, bounded post-processing operation, not an iterative optimizer. Accordingly, the theoretical properties of interest are stability, boundedness, and the absence of extrapolation, rather than asymptotic convergence. Because $w_{\min}\leq W_i \leq w_{\max}$, \eqref{eq:fusion} is a strict convex combination of two valid reconstructions. Therefore $x_{\mathrm{final},i}$ remains in the pixel-wise convex hull of $\mu_i$ and $x_i^*$ and cannot introduce a new value outside the range already spanned by the selected sample and the ensemble mean. Moreover,
\begin{equation}
\label{eq:residual_id}
x_{\mathrm{final},i}-\mu_i=(1-W_i)(x_i^*-\mu_i),
\end{equation}
and since $1-W_i\leq 1-w_{\min}=0.3$,
\begin{equation}
\label{eq:residual_bound}
\left|x_{\mathrm{final},i}-\mu_i\right|
\leq
0.3\,\left|x_i^*-\mu_i\right|.
\end{equation}
At the image level, this gives $\|\mathbf{x}_{\mathrm{final}}-\boldsymbol{\mu}\|_{\infty}\leq 0.3\,\|\mathbf{x}^*-\boldsymbol{\mu}\|_{\infty}$, which shows precisely why the fusion can simultaneously reduce randomness in uncertain regions and preserve detail in stable regions without introducing a new structural offset larger than the disagreement already present in the ensemble.

\subsection{Overall Algorithm}
Algorithm~\ref{alg:UGNS} delineates the UGNS pipeline, explicitly detailing the log inversion, extraction of the structural prior, the adaptive range-null diffusion steps, and final uncertainty-guided fusion.

\begin{algorithm}[!t]
\caption{\textbf{UGNS}: Uncertainty-Guided Null-Space Diffusion for Ultrasound Speckle Reduction}
\label{alg:UGNS}
\begin{algorithmic}[1]
\REQUIRE Observed B-mode $\mathbf{y} \in \mathbb{R}^{H \times W}$, wavelet operator $\mathcal{A}$, pretrained noise predictor $\boldsymbol{\epsilon}_\theta$
\REQUIRE Steps $T=100$, ensemble $N=10$, dynamic range $D=60\,$dB, mask center $c=0.4$, steepness $\gamma=3.5$
\ENSURE Reconstructed Image $\mathbf{x}_{\mathrm{final}}$
\STAGE{Stage 1: Physically-Consistent Structural Prior}
\STATE $\mathbf{y}_{\mathrm{lin}} \leftarrow 10^{(D \cdot \mathbf{y} - D)/20}$ \COMMENT{Log Inversion}
\STATE $\mathbf{E} \leftarrow \mathcal{A}(\mathrm{SRAD}(\mathbf{y}_{\mathrm{lin}}))$ \COMMENT{Structural Prior Module}
\STATE $\mathbf{N} \leftarrow \mathrm{clip}\!\left((\mathbf{E}-c)/(1-c),\; 0,\; 1\right)$
\STATE $\mathbf{M} \leftarrow \mathbf{N}^\gamma \odot G_{\mathrm{otsu}}(\mathbf{N})$ \COMMENT{Adaptive Weight Mask}
\STAGE{Stage 2: Adaptive Range-Null Reconstruction}
\FOR{$n = 1$ \TO $N$}
    \STATE $\mathbf{x}_T^{(n)} \sim \mathcal{N}(\mathbf{0}, \mathbf{I})$
    \FOR{$t = T, \dots, 1$}
        \STATE $\boldsymbol{\epsilon}^{(n)} \leftarrow \boldsymbol{\epsilon}_\theta(\mathbf{x}_t^{(n)}, t)$ \COMMENT{Noise Prediction}
        \STATE $\mathbf{x}_{0|t} \leftarrow \frac{1}{\sqrt{\bar{\alpha}_t}}\left( \mathbf{x}_t^{(n)} - \sqrt{1 - \bar{\alpha}_t}\boldsymbol{\epsilon}^{(n)} \right)$ \COMMENT{Clean State Prediction}
        \STATE $\tilde{\mathbf{x}}_{0|t} \leftarrow \mathcal{A}^\dagger \mathcal{A} \mathbf{y}_{\mathrm{lin}} + (\mathbf{I} - \mathcal{A}^\dagger \mathcal{A}) \mathbf{x}_{0|t}$ \COMMENT{Range-Null Projection}
        \STATE $\hat{\mathbf{x}}_{0|t} \leftarrow \mathbf{M} \odot \tilde{\mathbf{x}}_{0|t} + (\mathbf{1} - \mathbf{M}) \odot \mathbf{x}_{0|t}$ \COMMENT{Adaptive Mixing}
        \STATE $\lambda_t \leftarrow \mathrm{CosineSchedule}(t)$ \COMMENT{See Eq.~\eqref{eq:cosine}}
        \STATE $\hat{\mathbf{x}}_{0|t}^{\mathrm{guided}} \leftarrow (1 - \lambda_t)\hat{\mathbf{x}}_{0|t} + \lambda_t \mathbf{y}_{\mathrm{lin}}$ \COMMENT{Measurement Guidance}
        \STATE $\mathbf{x}_{t-1}^{(n)} \leftarrow \sqrt{\bar{\alpha}_{t-1}} \hat{\mathbf{x}}_{0|t}^{\mathrm{guided}} + \sqrt{1 - \bar{\alpha}_{t-1}} \boldsymbol{\epsilon}^{(n)}$ \COMMENT{DDIM Update}
    \ENDFOR
    \STATE $\hat{\mathbf{x}}_0^{(n)} \leftarrow \mathbf{x}_0^{(n)}$
\ENDFOR
\STAGE{Stage 3: Uncertainty-Guided Fusion}
\STATE $\boldsymbol{\mu} \leftarrow \frac{1}{N}\sum_{n=1}^{N} \hat{\mathbf{x}}_0^{(n)}$ \COMMENT{Ensemble Mean}
\STATE $\mathbf{U} \leftarrow \mathrm{Normalize}\left(\sqrt{\frac{1}{N}\sum_{n=1}^N (\hat{\mathbf{x}}_0^{(n)} - \boldsymbol{\mu})^2}\right)$ \COMMENT{Uncertainty Estimation}
\STATE $n^* \leftarrow \operatorname*{arg\,max}_{n} Q_n$, \quad $\mathbf{x}^* \leftarrow \hat{\mathbf{x}}_0^{(n^*)}$ \COMMENT{Optimal Sample Selection}
\STATE $\mathbf{W} \leftarrow \mathrm{clip}(0.7 + 0.28\,\mathbf{U},\; 0.7,\; 0.98)$
\STATE $\mathbf{x}_{\mathrm{final}} \leftarrow \mathbf{W} \odot \boldsymbol{\mu} + (\mathbf{1} - \mathbf{W}) \odot \mathbf{x}^*$
\RETURN $\mathbf{x}_{\mathrm{final}}$
\end{algorithmic}
\end{algorithm}

\section{Experimental Setup and Conditions}
\label{sec:experiments}

\subsection{Pretraining Datasets}
To expose the diffusion model to diverse ultrasound textures and speckle statistics, we curated a dataset of 5,929 ultrasound images from both \emph{in vivo} and \emph{in vitro} acquisitions.
The dataset comprises four subsets: (i) both4000 (3,551 images) with cross-sectional and longitudinal carotid views, (ii) cross500 (1,012 images) focusing on carotid cross-sectional views, (iii) vitro6000 (824 images) acquired from a CIRS phantom for standardized speckle statistics, and (iv) an additional 542 uncurated ultrasound images collected from multiple devices to increase data diversity and improve generalization.
No clean reference labels, paired noisy-clean annotations, or task-specific supervision were required; the diffusion prior is trained in a fully \emph{unsupervised generative} manner on raw unlabeled ultrasound images using a standard DDPM noise-prediction objective. This is distinct from self-supervised CNN approaches (e.g., Noise2Noise, Noise2Void) that rely on task-specific paired noisy images or image-specific blind-spot pretext tasks, and is consistent with the \emph{label-free} paradigm of the proposed framework.
All images were resized to $256 \times 256$ using Lanczos interpolation to preserve fine texture. We used single-channel grayscale inputs and normalized pixel intensities to $[-1, 1]$, which is the standard input range for diffusion models.

\subsection{Validation Datasets}
In our experimental evaluation, two widely used evaluation protocols were adopted: the PICMUS benchmark \cite{liebgott2016plane} and the DRUS/DRUSvar evaluation framework \cite{zhang2023reconstruction,zhang2024variance}. When evaluating the proposed method for comparative purposes, we strictly followed the implementation guidelines (e.g., data partition, test settings, and evaluation metrics) listed in the respective published papers for the aforementioned evaluation protocols \cite{zhang2023reconstruction,zhang2024variance,liebgott2016plane}; this guarantees the stability of comparative results.

The PICMUS benchmark \cite{liebgott2016plane} comprises the simulated contrast and resolution datasets (SC/SR), the experimental contrast and resolution datasets (EC/ER), and the \emph{in vivo} carotid cross-sectional and longitudinal datasets (CC/CL). The EC/ER datasets were experimentally acquired using a CIRS Model 040GSE phantom, and the corresponding RF channel data were recorded with a 128-element L11-4v linear-array transducer at a sampling rate of $20.832$\,MHz, with a transmit fractional bandwidth of $67\%$ and 75 steered plane waves spanning $-16^{\circ}$ to $+16^{\circ}$. Throughout all experiments, contrast and background quality were evaluated using the contrast-to-noise ratio (CNR), generalized contrast-to-noise ratio (gCNR) \cite{rodriguez2020gcnr}, and signal-to-noise ratio (SNR), while spatial resolution was assessed using the full width at half maximum (FWHM) of point targets. Also note that our evaluation study is not limited to a particular acquisition setting. Kindly note that our evaluation study is an \emph{extended} version of the standard PICMUS acquisition configuration. PICMUS uses a Verasonics Vantage system with an L11-4v probe at $5.208$\,MHz, with a 1PW input and a 75PWs compounded DAS reference, whereas the additional \emph{in vivo} carotid dataset uses a TPAC Pioneer system with an L11-5 probe at $5.0$\,MHz, with a 1PW input and a 65PWs compounding configurations.

\subsection{Diffusion Model Architecture}
We used the UNet backbone from guided diffusion \cite{dhariwal2021diffusion}.
The base channel width is 256 with channel multipliers $(1, 1, 2, 2, 4, 4)$. Multihead attention was applied at feature resolutions of $32 \times 32$, $16 \times 16$, and $8 \times 8$. We set the diffusion process length to $T=1{,}000$ steps and used a linear noise schedule with $\beta_t$ increasing from $10^{-4}$ to $0.02$.

\subsection{Implementation Details}
All models were implemented in PyTorch and all experiments were 
run on a single NVIDIA GeForce RTX 4090 GPU. For pretraining, we used 
AdamW with a learning rate $1 \times 10^{-4}$, batch size 8, and 
100 training epochs. 
During inference, UGNS uses a 100-step DDIM sampler with the 
Karras timestep schedule. For measurement guidance, we invert the 
log compression via $g^{-1}(\cdot)$ with dynamic range $D=60$\,dB 
to obtain the stabilized positive-envelope proxy 
$\mathbf{y}_{\mathrm{lin}}$, and use a cosine schedule for guidance 
strength ($\lambda_{\mathrm{start}}=0.3$ to $\lambda_{\mathrm{end}}=0.0$). 
The sym4 wavelet (level 3) was used for range-null decomposition. 
To obtain a stable envelope estimate, SRAD was applied for 15 
iterations. Masking was performed using $c=0.4$, $\gamma=3.5$, and 
an Otsu-based soft background gate. An ensemble size of $N=10$ was 
used for uncertainty-guided fusion, matching the setup of DRUSvar 
\cite{zhang2024variance} for a fair comparison.

\begin{table}[!t]
\centering
\caption{Inference runtime and peak GPU memory on $256\times256$
inputs. UGNS is measured on a single NVIDIA GeForce RTX 4090 GPU.
$^\dagger$DRUS requires a one-time SVD preprocessing
(${\approx}3$\,h, ${\approx}34$\,GB CPU RAM); the inference time
shown is a hardware-aware estimate (NVIDIA GeForce RTX~4090).}
\label{tab:runtime}
\resizebox{\columnwidth}{!}{%
\begin{tabular}{lccc}
\toprule
\textbf{Metric}
  & \textbf{UGNS ($N=1$)}
  & \textbf{UGNS ($N=10$)}
  & \textbf{DRUS ($N=10$)$^\dagger$} \\
\midrule
Parameters [M]           & $552.8$            & $552.8$            & $552.8$             \\
Diffusion steps ($T$)    & $100$ (DDIM)       & $100$ (DDIM)       & $50$ (DDRM)         \\
Inference time [s/image] & ${\approx}6.0$     & ${\approx}54$      & ${\approx}90$       \\
Peak GPU memory [MB]     & ${\approx}2{,}705$ & ${\approx}7{,}891$ & ${\approx}7{,}893$  \\
\bottomrule
\end{tabular}%
}
\end{table}

\begin{table*}[t!]
    \centering
    \caption{Quantitative comparison of PICMUS datasets. The best results are highlighted in \textbf{bold}, while the \underline{second-best results are underlined}. For resolution metrics (FWHM), lower values indicate better performance ($\downarrow$); for contrast metrics (gCNR, CNR[dB], SNR), higher values are better ($\uparrow$). Asterisks ($*$) indicate endpoints for which UGNS achieved the best value and was statistically superior to all competing methods after Bonferroni correction across 110 UGNS-to-comparator tests (11 endpoints $\times$ 10 comparators; prespecified one-sided paired $t$-test; $\alpha_{\mathrm{corrected}}=0.000455$). Statistics are computed from raw per-sample values; the table shows rounded values.}
    \label{tab:contrast_resolution}
    \resizebox{\textwidth}{!}{%
    \begin{tabular}{llllllllllll}
        \toprule
        \multirow{2}{*}{\textbf{Method}}
        & \multicolumn{3}{c}{\textbf{Exp. Contrast (EC)}}
        & \multicolumn{3}{c}{\textbf{Sim. Contrast (SC)}}
        & \multicolumn{3}{c}{\textbf{Exp. Resolution (ER)}}
        & \multicolumn{2}{c}{\textbf{Sim. Resolution (SR)}} \\
        \cmidrule(lr){2-4} \cmidrule(lr){5-7}
        \cmidrule(lr){8-10} \cmidrule(lr){11-12}
        & gCNR$\uparrow$ & CNR[dB]$\uparrow$ & SNR$\uparrow$
        & gCNR$\uparrow$ & CNR[dB]$\uparrow$ & SNR$\uparrow$
        & gCNR$\uparrow$ & FWHM\textsubscript{A}$\downarrow$
        & FWHM\textsubscript{L}$\downarrow$
        & FWHM\textsubscript{A}$\downarrow$
        & FWHM\textsubscript{L}$\downarrow$ \\
        \midrule
        DAS (1PW)~\cite{perrot2021think}
        & 0.5628 & \;\;\;2.553 & 3.307
        & 0.6436 & \;\;\;5.072 & 3.497
        & 0.5136 & \;\;0.1871 & \;\;0.6008
        & \;\;0.5319 & \;\;1.734 \\

        DAS (11PWs)~\cite{perrot2021think}
        & 0.6550 & \;\;\;5.483 & 3.156
        & 0.7317 & \;\;\;7.142 & 3.491
        & 0.5562 & \;\;0.1477 & \;\;\underline{0.3349}
        & \;\;0.5319 & \;\;0.6993 \\

        DAS (75PWs)~\cite{perrot2021think}
        & 0.7969 & \;\;\;8.059 & 3.211
        & 0.9058 & \;\;\;11.39 & 3.516
        & 0.5489 & \;\;0.1379 & \;\;0.3447
        & \;\;0.5319 & \;\;0.7584 \\

        \midrule
        Noise2Noise~\cite{lehtinen2018noise2noise}
        & 0.5393 & \;\;\;1.156 & 5.711
        & 0.6578 & \;\;\;4.599 & 5.338
        & 0.5300 & \;\;0.1970 & \;\;0.6008
        & \;\;0.5713 & \;\;1.655 \\

        Noise2Void~\cite{krull2019noise2void}
        & 0.5915 & \;\;\;4.206 & 3.852
        & 0.7014 & \;\;\;6.885 & 4.291
        & 0.5662 & \;\;0.3743 & \;\;0.5811
        & \;\;\underline{0.4925} & \;\;1.399 \\

        DENOmean~\cite{asgariandehkordi2023deep}
        & 0.7906 & \;\;\;8.200 & 3.002
        & 0.8893 & \;\;\;11.46 & 8.353
        & 0.8952 & \;\;0.1970 & \;\;0.5516
        & \;\;0.5811 & \;\;0.6796 \\

        DENOvar~\cite{asgariandehkordi2023deep}
        & 0.8474 & \;\;\;9.545 & 4.081
        & 0.9096 & \;\;\;12.25 & \textbf{16.33}
        & 0.9292 & \;\;0.2265 & \;\;0.4137
        & \;\;0.8372 & \;\;0.8471 \\

        DRUSmean~\cite{zhang2024variance}
        & 0.9011 & \;\;\;10.96 & 3.258
        & \underline{0.9477} & \;\;\;\underline{13.53} & 9.385
        & 0.9292 & \;\;0.1000 & \;\;0.5023
        & \;\;0.5910 & \;\;\textbf{0.5713} \\

        DRUSvar~\cite{zhang2024variance}
        & \underline{0.9826} & \;\;\;\textbf{15.35}
        & \underline{5.957}
        & 0.9089 & \;\;\;13.44 & \underline{16.23}
        & \underline{0.9474} & \;\;\textbf{0.0886}
        & \;\;\underline{0.3349}
        & \;\;0.5516 & \;\;0.8159 \\

        SDPS~\cite{stevens2025semantic}
        & 0.7303 & \;\;\;6.631 & 5.920
        & 0.8072 & \;\;\;8.757 & 6.645
        & 0.7450 & \;\;0.2955 & \;\;0.4531
        & \;\;0.5811 & \;\;2.935 \\

        \midrule
        \textbf{UGNS (Ours)}
        & \textbf{0.9948}$^{*}$ & \;\;\;\underline{11.29}
        & \textbf{8.155}$^{*}$
        & \textbf{0.9933}$^{*}$ & \;\;\;\textbf{19.81}$^{*}$
        & 12.83
        & \textbf{0.9512}$^{*}$ & \;\;\underline{0.0985}
        & \;\;\textbf{0.3053}$^{*}$
        & \;\;\textbf{0.4826}$^{*}$ & \;\;\underline{0.6107} \\
        \bottomrule
    \end{tabular}%
    }
\end{table*}

Table~\ref{tab:runtime} reports the inference cost and peak GPU 
memory for diffusion-based methods. As a lightweight analytical 
reference, SRAD runs at approximately $7.8$\,ms per image on CPU 
(15 iterations, $256 \times 256$ input, 20-run average). 
The increased runtime of the default ensemble setting ($N=10$) over 
the single-sample setting ($N=1$) reflects the tenfold growth in 
tensor computation per DDIM step rather than sequential repetition, 
since all ten samples are processed as a single batch. DRUS 
additionally requires a one-time system-matrix construction and SVD 
decomposition prior to inference (${\approx}3$\,h on Intel Core 
i9-14900K), with approximately $34$\,GB CPU RAM for SVD 
matrices; the per-image inference time of ${\approx}90$\,s shown in 
Table~\ref{tab:runtime} is a hardware-aware engineering estimate 
rather than a direct measurement under a unified protocol. 
Noise2Noise and Noise2Void require target-task retraining before 
inference; their per-image cost thereafter is a single CNN forward 
pass, expected to be substantially faster than iterative diffusion 
sampling but not formally benchmarked here. Formal wall-clock 
comparisons under a unified hardware protocol remain an important 
direction for future work.

\subsection{Evaluation Metrics}
\subsubsection{Resolution}
FWHM of point targets was assessed at $-6$\,dB. We report axial (FWHM\textsubscript{A}) and lateral (FWHM\textsubscript{L}) values for targets in the SR and ER datasets; lower values indicate better resolution.

\subsubsection{Contrast}
For contrast evaluation, we report the CNR and the generalized contrast-to-noise ratio (gCNR)
\cite{rodriguez2020gcnr}:
\begin{equation}
\mathrm{CNR} = 20 \log_{10}\left( \frac{|\mu_{\mathrm{in}} - \mu_{\mathrm{out}}|}{\sqrt{(\sigma_{\mathrm{in}}^2 + \sigma_{\mathrm{out}}^2) / 2}} \right)
\end{equation}
\begin{equation}
\mathrm{gCNR} = 1 - \int_{-\infty}^{\infty} \min \{ p_{\mathrm{in}}(v), p_{\mathrm{out}}(v) \}\,\mathrm{d}v
\end{equation}
CNR indicates contrast relative to noise, whereas gCNR quantifies histogram overlap (values closer to 1 indicate better separation).

\subsubsection{Background Quality}
To assess noise suppression, SNR was computed as
\begin{equation}
\label{eq:snr}
\mathrm{SNR} = \frac{\mu_{\mathrm{ROI}}}{\sigma_{\mathrm{ROI}}}
\end{equation}
where $\mu_{\mathrm{ROI}}$ and $\sigma_{\mathrm{ROI}}$ denote the mean and SD
computed within a homogeneous region of interest (ROI). A higher SNR indicates a
smoother background with reduced speckle.

\subsubsection{Statistical Verification}
The PICMUS phantom comparisons are performed using paired \emph{one-sided}
$t$\emph{-tests} because the direction of improvement is predefined by each
metric. Because higher values indicate better quality for contrast metrics
(gCNR, CNR[dB], SNR), the paired difference was defined as
$\Delta=m_{\mathrm{UGNS}}-m_{\mathrm{baseline}}$ and tested with
$H_1:\Delta>0$. Because lower values indicate better resolution for resolution
metrics (FWHM\textsubscript{A}, FWHM\textsubscript{L}), we defined
$\Delta=m_{\mathrm{baseline}}-m_{\mathrm{UGNS}}$, which again makes
$H_1:\Delta>0$ correspond to improved performance by UGNS. As this direction is
determined by the metric convention, rather than by post-hoc inspection of the
results, this approach is appropriate for the prespecified superiority
hypothesis.

For UGNS, the ensemble size was $N=10$. Each sample was generated from the same
trained prior using a distinct initial Gaussian noise vector
$\mathbf{x}_T\sim\mathcal{N}(\mathbf{0},\mathbf{I})$, yielding $n=10$ paired per-sample
metric values for the ensemble-based statistical analysis. For statistically
significant comparisons, we report 95\% confidence intervals (CIs) of the paired
mean difference $\Delta$, computed from the raw paired per-sample differences
$\{\Delta_i\}_{i=1}^{10}$ rather than by subtracting the rounded aggregate values
shown in Table~\ref{tab:contrast_resolution}. This approach is consistent with
the DRUSvar framework~\cite{zhang2024variance}, which also interprets sampling
dispersion as an informative quality indicator.

For the higher-is-better SC phantom gCNR metric, UGNS achieved a mean of
$0.9933$ compared with $0.9477$ for DRUSmean, yielding an effect size of
$\Delta_{\mathrm{gCNR}}=0.0456$ (95\% CI: $[0.0445,\,0.0467]$; one-sided paired
$t$-test: $t$-statistic $=95.4$, $\mathrm{df}=9$, $p<0.0001$). For the
lower-is-better ER phantom lateral resolution metric, UGNS achieved
FWHM\textsubscript{L} $=0.3053$\,mm compared with $0.5023$\,mm for DRUSmean.
Using $\Delta=m_{\mathrm{DRUSmean}}-m_{\mathrm{UGNS}}$, the unstandardized
effect size was $\Delta_{\mathrm{FWHM_L}}=0.1970$\,mm (95\% CI: $[0.1385,\,0.2555]$\,mm;
one-sided paired $t$-test: $t$-statistic $=7.62$, $\mathrm{df}=9$, $p<0.0001$).

\begin{figure*}[t!]
\centering
\includegraphics[width=0.95\linewidth]{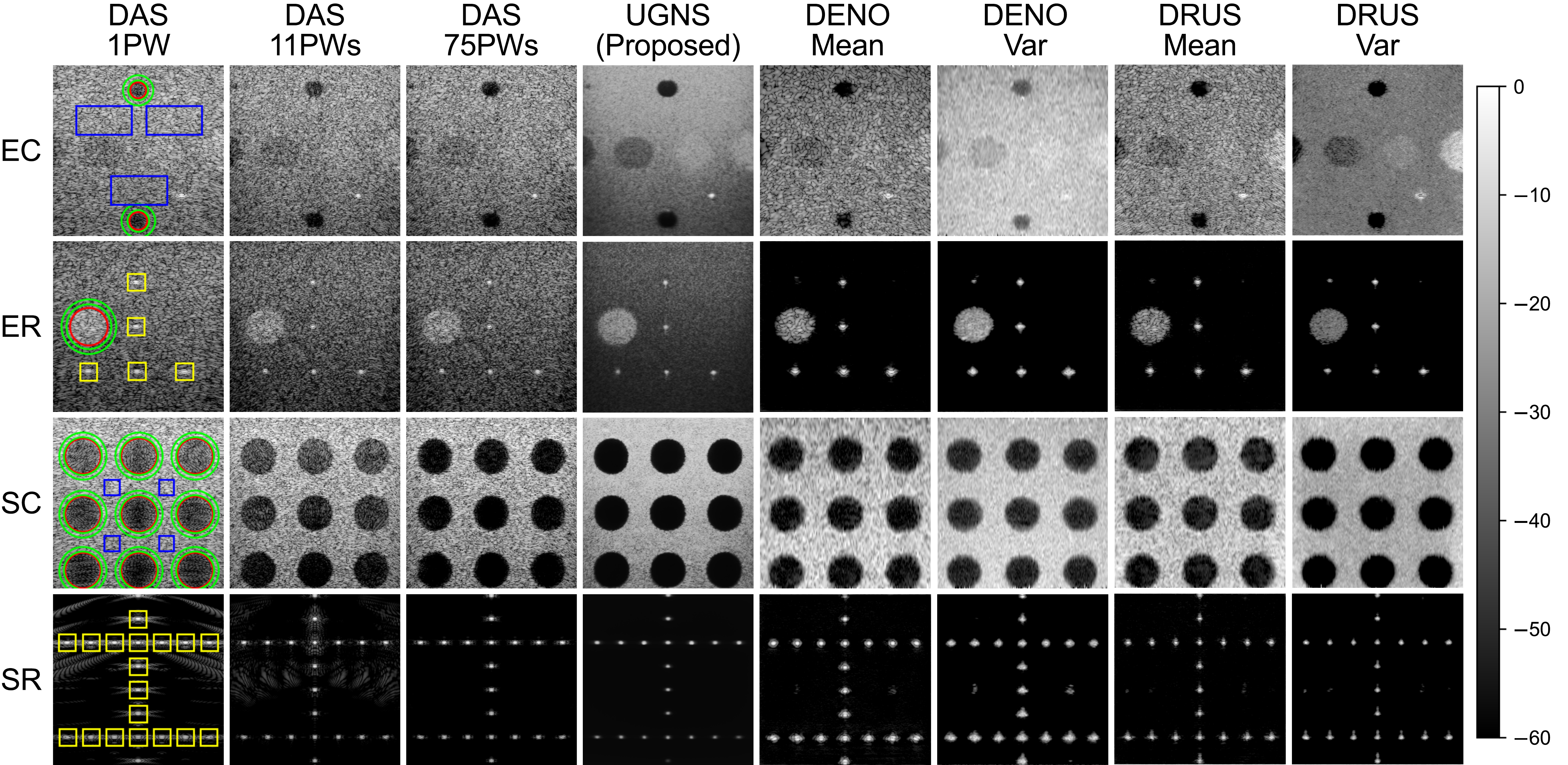}
\caption{\textbf{Visual comparison of PICMUS phantom datasets.} From top to bottom: experimental contrast (EC), experimental resolution (ER), simulated contrast (SC), and simulated resolution (SR). UGNS (fourth column) preserves target structures and maintains a natural background texture without the heavy speckle of DAS or the over-smoothing of DRUS variants.}
\label{fig:figure2}
\end{figure*}

To control the family-wise error rate, Bonferroni correction was applied across
the complete family of 110 UGNS-to-comparator tests (11 endpoints $\times$ 10
comparators), yielding
$\alpha_{\mathrm{corrected}}=0.05/110=0.000455$. Asterisks in
Table~\ref{tab:contrast_resolution} indicate comparisons whose raw one-sided
$p$-values remained below this corrected threshold (equivalently,
Bonferroni-adjusted $p<0.05$), and thus remain statistically significant after
correction.

\section{Results}
\label{sec:results}
\begin{figure*}[t!]
\centering
\includegraphics[width=0.95\linewidth]{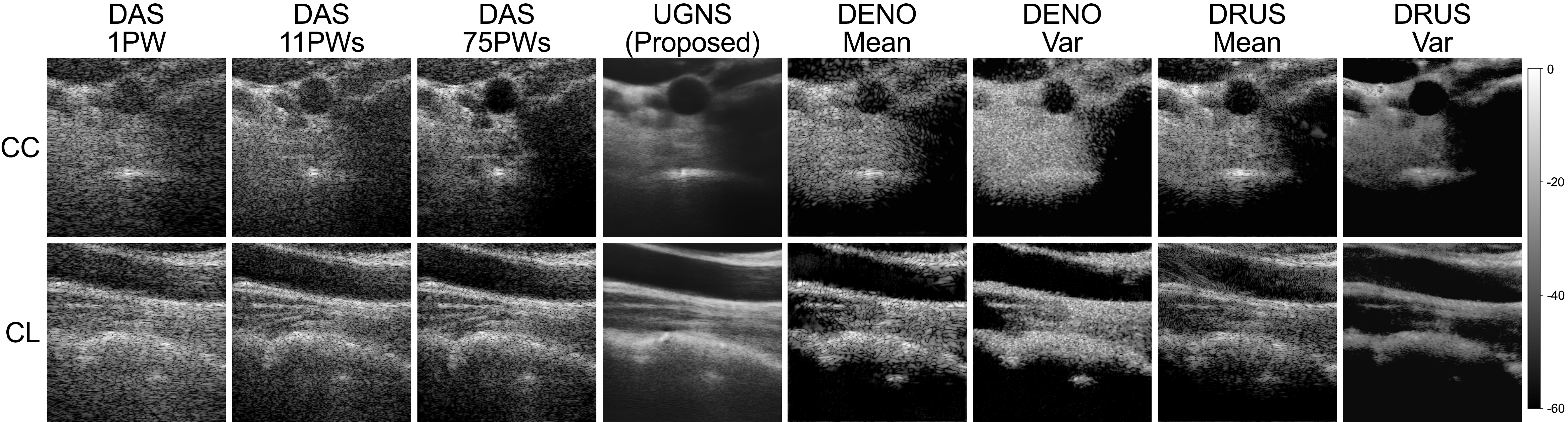}
\caption{\textbf{Qualitative comparison of \emph{in vivo} carotid data.} The top row presents cross-sectional views (CC) and the bottom row shows longitudinal views (CL). DAS (1PW/11PWs) retains residual speckle, whereas the DRUS variants tend to oversmooth tissue details. UGNS reconstructs a clean vessel lumen while preserving the sharp boundaries of the vessel wall and surrounding tissues.}
\label{fig:figure3}
\end{figure*}

We report quantitative and qualitative results on the PICMUS benchmark (EC, SC, ER, and SR) and \emph{in vivo} carotid datasets to evaluate our UGNS. Comparisons are made with conventional DAS beamforming and recent diffusion-based baselines, including DENO \cite{asgariandehkordi2023deep}, DRUS (mean/var) \cite{zhang2023reconstruction} and the semantic-guided diffusion method SDPS \cite{stevens2025semantic}, as well as self-supervised reference methods (Noise2Noise \cite{lehtinen2018noise2noise} and Noise2Void \cite{krull2019noise2void}).

\subsection{Quantitative Analysis}
Table~\ref{tab:contrast_resolution} presents the contrast (gCNR, CNR, and SNR) and resolution (FWHM\textsubscript{A}/FWHM\textsubscript{L}) metrics for comparison with other state-of-the-art (SOTA) methods. In particular, semantic-guided diffusion posterior sampling (SDPS) \cite{stevens2025semantic} SOTA method was compared because it is a recently developed label-free ultrasound restoration method that uses a semantic segmentation map.
For the contrast datasets, UGNS attained the highest gCNR on both EC and SC.
On EC, UGNS increased gCNR from 0.7969 (DAS, 75PWs) to 0.9948 (+24.8\% relative),
achieved the highest SNR (8.155), and yielded the second-highest CNR (11.29).
The highest SNR values on SC were obtained by DENOvar (16.33) and DRUSvar (16.23),
which tended to over-suppress the background texture (Fig.~\ref{fig:figure2}).

For the resolution datasets, UGNS achieved the highest gCNR on ER (0.9512), the lowest raw per-sample mean lateral FWHM (FWHM\textsubscript{L} $=0.3053$\,mm), and the second-lowest axial FWHM (FWHM\textsubscript{A} $=0.0985$\,mm).
On SR, UGNS attained the lowest axial FWHM (FWHM\textsubscript{A} $=0.4826$\,mm) and the second-lowest lateral FWHM (FWHM\textsubscript{L} $=0.6107$\,mm), which was close to the best value from DRUSmean (0.5713\,mm). Overall, these findings demonstrate that UGNS provides a favorable contrast-resolution trade-off by preserving range-space tissue structure while suppressing speckle through null-space diffusion, although performance advantages vary across metrics and datasets.

Note that \textbf{clinical utility of the reported metric results in our quantitative analysis} is explained as follows. Higher gCNR is now interpreted as reduced overlap between the target and background intensity distributions and improved pixel-level separability for an ideal observer. Lower FWHM is interpreted as a narrower point-spread response and a greater capacity to delineate closely spaced boundaries. CNR and SNR are described as complementary measures of local target conspicuity and background stability. Because CNR and SNR may also increase under aggressive smoothing, we emphasize that these metrics should be interpreted jointly with gCNR, FWHM, and the qualitative preservation of tissue texture. Accordingly, such findings are presented as preliminary evidence of a favorable contrast-resolution trade-off rather than as evidence of improved diagnostic accuracy.

\subsection{Qualitative Analysis}
We further compare visual quality using PICMUS phantom data and \emph{in vivo} carotid images.

\subsubsection{Evaluation of the PICMUS Phantom Data}
Fig.~\ref{fig:figure2} presents the results of the four PICMUS datasets. On EC and SC, DAS produced blurred cyst boundaries due to speckle. DRUSvar largely removed background speckle but also erased characteristic texture, resulting in an overly ``binary'' appearance. In contrast, UGNS produced dark cyst interiors while preserving a realistic background 
texture, indicating that the proposed structural prior guides sampling without distorting anatomical structure.

\subsubsection{\emph{In Vivo} Evaluation}
Many recent ultrasound benchmark studies, including the 2024 EUSIPCO study~\cite{zhang2024variance}, have evaluated their methods primarily through qualitative \emph{in vivo} comparison. To complement this qualitative assessment with a more objective indicator, we performed an ROI-based local CNR analysis on two additional carotid \emph{in vivo} sequences. These sequences belong to the additional \emph{in vivo} dataset introduced in~\cite{zhang2023informative}, acquired from the carotid artery of a volunteer using a TPAC Pioneer ultrasound system equipped with an L11-5 probe, with a transmit pulse of $5.0$\,MHz center frequency and a fractional bandwidth of $50\%$. For this evaluation, the high-quality reference was obtained by coherent compounding of 65 plane waves (65PWs).

\begin{figure}[t!]
\centering
\includegraphics[width=\linewidth]{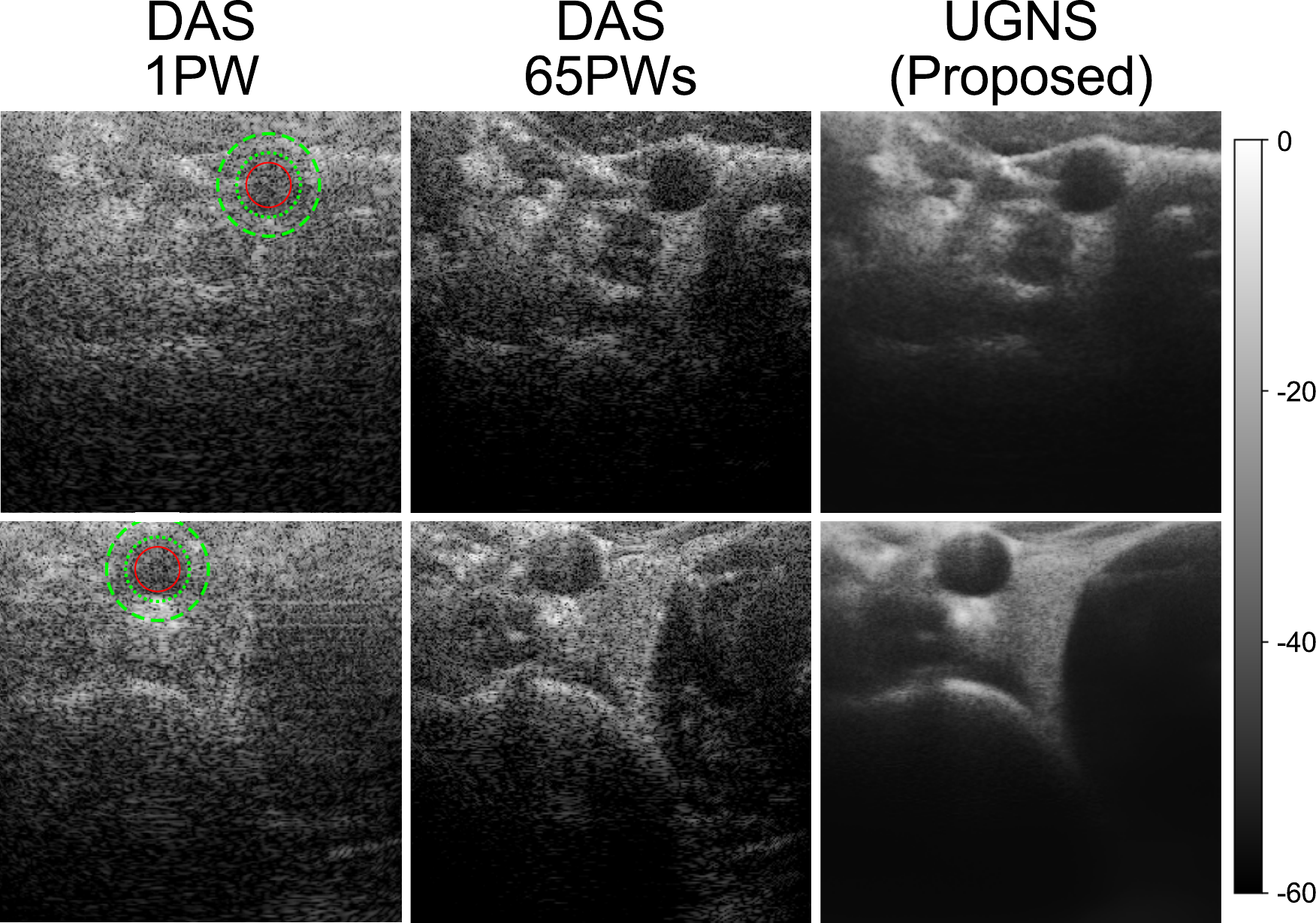}
\caption{\textbf{ROI placement for the \emph{in vivo} local CNR analysis.}
From left to right, the columns show DAS (1PW), DAS (65PWs), and UGNS (proposed, 1PW input).
From top to bottom, the rows show the cross-sectional (CC) and longitudinal (CL) carotid
sequences acquired using the TPAC Pioneer system with an L11-5 probe.
The lumen and vessel-wall ROIs were defined on the DAS~(65PWs) reference images and applied
identically to all methods. For clarity, the ROIs are overlaid on the DAS~(1PW) images:
the red solid circle denotes the lumen ROI, and the region between the two green dashed
circles denotes the annular vessel-wall ROI.}
\label{fig:figure4}
\end{figure}

Circular lumen-versus-wall ROIs were placed on the vessel lumen and the adjacent wall identified on the DAS~(65PWs) reference images, following the gCNR ROI convention, and the identical ROIs were applied to all compared methods (DAS~1PW, DAS~65PWs, and UGNS) for CNR computation (Table~\ref{tab:invivo_cnr}; ROI placement shown in Fig.~\ref{fig:figure4}).

On the \emph{in vivo} carotid images (Fig.~\ref{fig:figure3}), DRUS-based methods tended to oversmooth, blurring the vessel-wall boundaries and attenuating fine texture, while DAS retained residual speckle inside the vessel even with compounding, preventing a uniformly dark lumen. By applying consistency correction on the inverse-compressed positive-envelope proxy, UGNS suppressed intraluminal clutter while preserving the sharp vessel-wall boundaries and the texture of surrounding muscle. This is reflected in the local CNR values of Table~\ref{tab:invivo_cnr}: from a single plane wave, UGNS attained a mean of $8.48$\,dB (CC $8.17$\,dB, CL $8.80$\,dB), exceeding DAS (1PW, $1.36$\,dB) by $7.12$\,dB and the 65PWs compounded reference ($3.61$\,dB) by $4.87$\,dB, with the largest margin on the cross-sectional sequence where the DAS (1PW) lumen was clutter-dominated ($-0.88$\,dB). These values are reported as an objective companion to the qualitative comparison rather than as clinical validation.

\begin{table}[t!]
\centering
\caption{Local CNR (dB) on two \emph{in vivo} carotid sequences acquired with a TPAC Pioneer ultrasound system and an L11-5 probe. The red solid circle denotes the vessel lumen ROI (hypoechoic inner region) while the green dashed circle denotes the vessel wall ROI (hyperechoic annular region). Higher values are better. Values for UGNS use a single plane-wave input (1PW).}
\label{tab:invivo_cnr}
\resizebox{\columnwidth}{!}{%
\begin{tabular}{lccc}
\toprule
\textbf{Method} & \textbf{CC local CNR[dB]} & \textbf{CL local CNR[dB]} & \textbf{Mean[dB]} \\
\midrule
DAS (1PW)  & $-0.88$ & 3.60 & 1.36 \\
DAS (65PWs) & 4.37 & 2.85 & 3.61 \\
\textbf{UGNS (Ours, 1PW)} & \textbf{8.17} & \textbf{8.80} & \textbf{8.48} \\
\bottomrule
\end{tabular}%
}
\end{table}

\subsection{Ablation Study}
\label{sec:ablation}

\begin{table*}[!tp]
    \centering
    \caption{\textbf{Evaluating contribution of UGNS component on the PICMUS phantom benchmark.}
    Contrast metrics are higher-is-better ($\uparrow$), and FWHM values are
    lower-is-better ($\downarrow$). All non-Full variants follow the same ablation
    protocol, except for \emph{w/o uncertainty fusion}, which uses a single sample
    ($N{=}1$). FWHM values are intended for relative comparison within this table.}
    \label{tab:ablation_component}
    \setlength{\tabcolsep}{4pt}
    \scriptsize
    \begin{tabular}{lccccccccccc}
        \toprule
        \multirow{2}{*}{\textbf{Variant}}
        & \multicolumn{3}{c}{\textbf{EC}} & \multicolumn{3}{c}{\textbf{SC}}
        & \multicolumn{3}{c}{\textbf{ER}} & \multicolumn{2}{c}{\textbf{SR}} \\
        \cmidrule(lr){2-4}\cmidrule(lr){5-7}\cmidrule(lr){8-10}\cmidrule(lr){11-12}
        & gCNR$\uparrow$ & CNR[dB]$\uparrow$ & SNR$\uparrow$
        & gCNR$\uparrow$ & CNR[dB]$\uparrow$ & SNR$\uparrow$
        & gCNR$\uparrow$ & FWHM\textsubscript{A}$\downarrow$
        & FWHM\textsubscript{L}$\downarrow$
        & FWHM\textsubscript{A}$\downarrow$
        & FWHM\textsubscript{L}$\downarrow$ \\
        \midrule
        \textbf{Full (UGNS)} & 0.9948 & 11.29 & 8.155 & 0.9933 & 19.81 & 12.83
        & 0.9512 & 0.0985 & 0.3053 & 0.4826 & 0.6107 \\
        \midrule
        w/o Structural Prior (SRAD+Wavelet) & 0.9910 & 11.17 & 7.94 & 0.9963 & 18.15 & 10.73
        & 0.9400 & 0.3349 & 0.3349 & 0.5122 & 0.9653 \\
        w/o SRAD only & 0.9905 & 11.17 & 7.92 & 0.9970 & 17.78 & 10.01
        & 0.9387 & 0.3250 & 0.3349 & 0.5122 & 0.8743 \\
        w/o Wavelet only & 0.9910 & 11.17 & 7.94 & 0.9963 & 18.16 & 10.77
        & 0.9400 & 0.3349 & 0.3349 & 0.5122 & 0.9751 \\
        w/o range anchoring (pure diffusion) & 0.9905 & 11.14 & 7.85 & 0.9961 & 15.09 & 6.28
        & 0.9410 & 0.3546 & 0.3349 & 0.5910 & 0.8569 \\
        single-sample inference ($N{=}1$) & 0.8958 & 9.46 & 6.06 & 0.9930 & 16.56 & 7.23
        & 0.8424 & 0.1576 & 0.4334 & 0.4432 & 0.9850 \\
        mean-only fusion ($N{=}10$) & 0.9934 & 11.49 & 8.56 & 0.9967 & 17.73 & 9.91
        & 0.9728 & 0.3349 & 0.3940 & 0.5417 & 0.8569 \\
        \bottomrule
    \end{tabular}
\end{table*}

\begin{table*}[!tp]
    \centering
    \caption{\textbf{Exploring hyperparameter sensitivity of UGNS on the PICMUS phantom benchmark.}
    Each row changes a single hyperparameter around the selected configuration:
    $c{=}0.4$, $\gamma{=}3.5$, wavelet level 3, SRAD 15 iterations, $N{=}10$, and
    $\lambda_{\mathrm{start}}{=}0.3$. Metric conventions follow
    Table~\ref{tab:ablation_component}.}
    \label{tab:ablation_params}
    \setlength{\tabcolsep}{4pt}
    \scriptsize
    \begin{tabular}{lccccccccccc}
        \toprule
        \multirow{2}{*}{\textbf{Configuration}}
        & \multicolumn{3}{c}{\textbf{EC}} & \multicolumn{3}{c}{\textbf{SC}}
        & \multicolumn{3}{c}{\textbf{ER}} & \multicolumn{2}{c}{\textbf{SR}} \\
        \cmidrule(lr){2-4}\cmidrule(lr){5-7}\cmidrule(lr){8-10}\cmidrule(lr){11-12}
        & gCNR$\uparrow$ & CNR[dB]$\uparrow$ & SNR$\uparrow$
        & gCNR$\uparrow$ & CNR[dB]$\uparrow$ & SNR$\uparrow$
        & gCNR$\uparrow$ & FWHM\textsubscript{A}$\downarrow$
        & FWHM\textsubscript{L}$\downarrow$
        & FWHM\textsubscript{A}$\downarrow$
        & FWHM\textsubscript{L}$\downarrow$ \\
        \midrule
        \textbf{Full (UGNS)} & 0.9948 & 11.29 & 8.155 & 0.9933 & 19.81 & 12.83
        & 0.9512 & 0.0985 & 0.3053 & 0.4826 & 0.6107 \\
        \midrule
        mask $c{=}0.35$ & 0.9905 & 11.17 & 7.91 & 0.9967 & 17.63 & 9.56
        & 0.9389 & 0.3250 & 0.3349 & 0.5122 & 0.8743 \\
        mask $c{=}0.45$ & 0.9905 & 11.17 & 7.91 & 0.9961 & 17.51 & 9.50
        & 0.9389 & 0.3250 & 0.3349 & 0.5122 & 0.8743 \\
        mask $\gamma{=}3.0$ & 0.9862 & 11.20 & 7.86 & 0.9959 & 17.45 & 9.45
        & 0.9389 & 0.3250 & 0.3349 & 0.5122 & 0.8743 \\
        mask $\gamma{=}4.0$ & 0.9905 & 11.17 & 7.91 & 0.9969 & 17.56 & 9.35
        & 0.9389 & 0.3250 & 0.3349 & 0.5122 & 0.8743 \\
        ensemble $N{=}1$ & 0.8958 & 9.46 & 6.06 & 0.9930 & 16.56 & 7.23
        & 0.8424 & 0.1576 & 0.4334 & 0.4432 & 0.9850 \\
        ensemble $N{=}15$ & 0.9953 & 11.32 & 8.12 & 0.9961 & 17.59 & 9.68
        & 0.9482 & 0.3250 & 0.3349 & 0.5220 & 0.8644 \\
        wavelet level $2$ & 0.9905 & 11.18 & 7.93 & 0.9972 & 18.05 & 10.49
        & 0.9394 & 0.3250 & 0.3349 & 0.5319 & 0.9751 \\
        wavelet level $4$ & 0.9662 & 10.72 & 4.93 & 0.9860 & 14.85 & 6.05
        & 0.9021 & 0.2856 & 0.3349 & 0.5614 & 0.9357 \\
        SRAD iter $5$ & 0.9905 & 11.17 & 7.92 & 0.9968 & 17.79 & 9.87
        & 0.9389 & 0.3250 & 0.3349 & 0.5122 & 0.8743 \\
        SRAD iter $30$ & 0.9905 & 11.17 & 7.91 & 0.9959 & 17.28 & 9.10
        & 0.9389 & 0.3250 & 0.3349 & 0.5122 & 0.8743 \\
        $\lambda_{\mathrm{start}}{=}0.2$ & 0.9089 & 9.13 & 8.68 & 0.9947 & 16.61 & 10.35
        & 0.8642 & 0.5516 & 0.5811 & 0.5122 & 0.9062 \\
        $\lambda_{\mathrm{start}}{=}0.4$ & 0.9976 & 11.81 & 7.69 & 0.9962 & 17.34 & 8.67
        & 0.9362 & 0.3152 & 0.3349 & 0.5417 & 0.9751 \\
        \bottomrule
    \end{tabular}
\end{table*}

To examine the contribution of each UGNS component, we evaluated the full UGNS
model and the following variants on the PICMUS phantom benchmark (under a fixed
random seed and a common evaluation protocol): (1) w/o structural prior
(SRAD~+~wavelet), (2) w/o SRAD only, (3) w/o wavelet only, (4) w/o range
anchoring (pure diffusion), (5) w/o uncertainty fusion, and (6) mean-only
fusion. In each case, only the target component was changed, except for the
\emph{w/o uncertainty fusion} variant, which is a single-sample configuration by
construction. The Full row denotes the selected UGNS configuration, and the FWHM
values in Tables~\ref{tab:ablation_component}--\ref{tab:ablation_params} are used
for within-table comparison and are not directly interchangeable with the
per-sample-mean resolution convention used in
Table~\ref{tab:contrast_resolution}.

The results in Table~\ref{tab:ablation_component} show that uncertainty-guided
fusion and range anchoring are the most influential components. Removing
uncertainty-guided fusion reduces EC gCNR from $0.9948$ to $0.8958$ and ER gCNR
from $0.9512$ to $0.8424$. In addition, removing range anchoring also degrades
background-fidelity metrics, reducing SC CNR from $19.81$ to $15.09$\,dB and SC
SNR from $12.83$ to $6.28$. The structural prior has a smaller effect on the
saturated EC/SC gCNR values, but it affects the resolution trade-off, as shown by
the degradation of SR lateral FWHM from $0.6107$ to $0.9653$ when it is removed.
Mean-only fusion improves some contrast-oriented metrics but worsens several
FWHM values.

Moreover, further ablation study has been performed to investigate the effect of
hyperparameter sensitivity. Table~\ref{tab:ablation_params} reports restoration
performance with varying hyperparameters of the proposed UGNS model. Note that
the hyperparameters considered include the mask parameters $(c,\gamma)$, wavelet
decomposition level, SRAD iteration count, ensemble size $N$, and
guidance-start parameter $\lambda_{\mathrm{start}}$. The results show that the
tested parameter values generally preserve the high contrast metrics while
altering the contrast-resolution trade-off. For example, increasing the ensemble
size to $N{=}15$ slightly improves some EC contrast values without improving
overall resolution, whereas reducing it to $N{=}1$ substantially decreases
contrast robustness. Changing the wavelet decomposition level in either
direction also degrades both contrast and resolution relative to the selected
configuration.

\section{Discussion}
\label{sec:discussion}

\subsection{Model Independence and Runtime Performance}
We formulate ultrasound speckle reduction using the additive 
observation model defined in Eq. \eqref{eq:add_model}, which holds in the 
\emph{normalized log-compressed domain} after homomorphic 
transformation of the multiplicative envelope model 
\eqref{eq:mult_model}, consistent with classical statistical 
acoustics. DDRM-style approaches (including DRUS) require an 
explicit system matrix $\mathbf{H}$ (i.e., 
$\mathbf{y}=\mathbf{H}\mathbf{x}+\mathbf{n}$) and its SVD to 
enforce consistency. This dependency introduces practical 
limitations, as $\mathbf{H}$ depends on the transducer geometry, 
center frequency, sampling rate, and beamforming settings. Changes 
in device configuration can require recomputing a computationally 
expensive SVD, potentially leading to model mismatch.

In contrast, UGNS adopts a signal-driven formulation that avoids $\mathbf{H}$ and SVD at inference time. By using a discrete wavelet 
transform to define the range-null decomposition, UGNS separates 
tissue structure from speckle using spatial-frequency 
characteristics alone. This system-matrix-free inference design, 
together with source pretraining followed by target deployment without 
task-specific fine-tuning, is the main model-independence advantage 
supported by the current study. The current evidence supports partial cross-dataset generalization on the PICMUS benchmark and \emph{in vivo} carotid data. Establishing broader hardware-independent generalization will require cross-device, cross-frequency, and cross-probe validation on data acquired from different transducers and acquisition setups.

As reported in Table~\ref{tab:runtime}, our UGNS diffusion backbone contains
$552.8$M parameters with the default 100-step DDIM sampler. For inference time,
approximately $6.0$\,s per image for $N=1$ and approximately $54$\,s per image
for the $N=10$ ensemble are required on a single NVIDIA GeForce RTX~4090, with
peak GPU memory usage of approximately $2.7$\,GB and $7.9$\,GB, respectively.
Note that the target application of our UGNS framework is offline,
non-time-critical applications, such as retrospective enhancement or
single-frame plane-wave reconstruction, rather than real-time clinical imaging.
Further acceleration may be achieved through faster diffusion
samplers~\cite{lu2022dpmsolver} and model-compression
techniques~\cite{li2023qdiffusion}. These strategies could reduce the number of
reverse diffusion steps, memory usage, and per-frame latency while preserving
the overall UGNS framework.

\subsection{Stability of Linear-Domain Processing and the Role of the Inverse-Compressed Proxy}
The domain used for range-null decomposition and guidance strongly influences restoration quality. Because log compression is nonlinear and amplifies low-intensity background noise, directly enforcing consistency in the log-compressed domain can disrupt the approximate additivity that would otherwise hold after log compression of the positive envelope. UGNS addresses this by performing consistency correction on a \emph{stabilized positive-envelope proxy} obtained via inverse log compression $\mathbf{y}_{\mathrm{lin}}=g^{-1}(\mathbf{y})$, while the diffusion prior itself operates in the normalized log-compressed domain. We do not claim that the raw envelope satisfies an additive relation; rather, the additive relation $\mathbf{y}=\mathbf{x}+\mathbf{n}+\boldsymbol{\eta}_{\mathrm{sys}}$ is used only in the log-compressed domain, and $\mathbf{y}_{\mathrm{lin}}$ serves as a model-consistent reference image for consistency correction---not as a claim of equality with the absolute acoustic pressure. Acoustic distortion and sound-speed inhomogeneity are absorbed into $\boldsymbol{\eta}_{\mathrm{sys}}$, which admits the element-wise upper bound stated in Appendix~E under bounded multiplicative system distortion. This makes the residual modeling error explicit rather than hidden. UGNS also avoids the ground truth paradox that limits supervised models because consistency is enforced against the observed proxy, rather than against paired clean labels.

\subsection{Clinical Fidelity and Contrast-Resolution Trade-off}
Variance-based strategies such as DRUSvar can achieve high numerical contrast metrics by aggressively suppressing low-intensity background regions; however, the outputs may appear unnaturally ``clean'' and deviate from standard B-mode appearance. Using heavily compounded DAS (75PWs) as a visual reference, UGNS suppresses clutter while retaining a natural texture profile, striking a more balanced contrast-resolution trade-off than the aggressive background suppression observed in DRUSvar. In several settings, UGNS approaches 75PWs-quality imaging from a single plane-wave input. However, we do not claim that UGNS uniformly rivals 75PWs across all criteria, nor that metric-level gains establish clinical equivalence. Larger cohorts, blinded expert assessment, and comparisons with supervised and self-supervised baselines are needed to support any stronger claims.

\subsection{Limitations and Future Work}
The proposed method has the following limitations. First, the wavelet-based
range-null decomposition provides an approximate frequency-based separation
rather than an exact physical separation between tissue structure and speckle,
which may induce the partial attenuation of fine anatomical structures
containing high-frequency components. Second, the ensemble-derived uncertainty
represents empirical sampling dispersion and should not be interpreted as a
calibrated probability of reconstruction error or diagnostic uncertainty.
Finally, the current validation using the PICMUS benchmark and two carotid
sequences does not establish generalization across different organs,
pathologies, devices, or acquisition settings.

To address these limitations, future research work will include (1) the
improvement of adaptive range-null decomposition and calibrated uncertainty
estimation, (2) more comprehensive validation including large-scale patient
cohorts, diverse scanners and probes, different anatomical regions, and
institutional acquisition protocols, and (3) employment of faster diffusion
samplers and model-compression techniques for reducing the number of reverse
diffusion steps, memory consumption, and per-frame latency.

\section{Conclusion}
\label{sec:conclusion}

This study proposes UGNS, a label-free generative framework for 
ultrasound speckle reduction. By enforcing consistency correction 
on a stabilized positive-envelope proxy and utilizing the discrete wavelet transform as 
a system-matrix-free linear operator, UGNS bypasses computationally 
expensive acoustic system modeling and effectively avoids the ground 
truth paradox that hinders existing supervised models. The proposed method integrates an adaptive range-null-space reconstruction to 
preserve deterministic tissue morphology and an uncertainty-guided 
fusion mechanism to suppress stochastic speckle noise. Extensive 
evaluations using the PICMUS benchmark and \emph{in vivo} datasets 
demonstrate that UGNS achieves a favorable contrast-resolution 
trade-off, consistently yielding competitive gCNR across 
diverse datasets. Notably, UGNS produces reconstructions that 
approach the visual quality of heavily compounded DAS (75PWs) from 
a single 1PW acquisition, offering a label-free solution 
for ultrasound image enhancement without the requirement for 
large-scale paired datasets or device-specific system modeling.

\appendix[Theoretical Justification of the Linear-Domain Signal Model
and Upper Bounds on the Residual]
\label{appendix:upper_bound}

This appendix provides the formal derivation supporting the
signal-model claims in Section~III-A. It includes:
(A)~the starting acoustic model,
(B)~the exactness of log-domain linearization,
(C)~the role of the discrete-wavelet-transform-based range-null operator,
(D)~the stability of the linear assumption under acoustic distortion
and sound-speed inhomogeneity, and
(E)~the explicit upper bounds on the residual between the UGNS signal
model and the true ultrasound envelope.

\subsection*{A. Starting from Pulse--Echo Acoustics}

Under the first-order Born approximation, the received RF echo
from a weakly scattering medium is modeled as:
\begin{equation}
  r(\mathbf{r}) = \int h(\mathbf{r}-\mathbf{r}')\,
                  f(\mathbf{r}')\,\mathrm{d}\mathbf{r}',
\end{equation}
where $f(\mathbf{r})$ is tissue reflectivity, $h(\cdot)$ is the
pulse--echo point-spread function (PSF), and $\mathbf{r}$ denotes
spatial coordinates. After coherent compounding and envelope
detection, the positive envelope is:
\begin{equation}
  \mathbf{i} = \mathbf{s} \odot \mathbf{u},
  \label{eq:app_mult}
\end{equation}
where $\mathbf{s} = |h \ast f|$ is the deterministic structural
envelope (band-limited tissue image through the PSF) and
$\mathbf{u} \in \mathbb{R}_{+}^{H\times W}$ is a strictly
positive, spatially correlated stochastic speckle field.
Under fully developed speckle, each pixel of $\mathbf{u}$
follows a Rayleigh distribution~\cite{burckhardt1978speckle,
wagner1983statistics}.

\subsection*{B. Exactness of the Log-Domain Linearization}

Applying the normalized log-compression operator
$g(\cdot)$ with a dynamic range $D = 60\,$dB to
\eqref{eq:app_mult} yields the \emph{exact} additive identity:
\begin{equation}
  \mathbf{y} = \mathbf{x} + \mathbf{n},
  \label{eq:app_exact}
\end{equation}
where $\mathbf{y} = g(\mathbf{i})$, $\mathbf{x} = g(\mathbf{s})$,
and $\mathbf{n} = g(\mathbf{u}) - g(\mathbf{1})$ is the
log-speckle additive noise term.
This is the classical homomorphic reformulation
\cite{oppenheim1968nonlinear,achim2001novel};
no approximation is introduced at this step.
Numerical stability is maintained by clipping $\mathbf{y},
\mathbf{x} \in [0,1]$ and using a small constant $\varepsilon > 0$
inside $g(\cdot)$. In particular, for $\mathbf{y}\in[0,1]$ and $D=60\,$dB, the 
inverse-compressed proxy satisfies 
$\mathbf{y}_{\mathrm{lin}}=g^{-1}(\mathbf{y})\in(10^{-3},1]$, so it 
remains on the same normalized $[0,1]$ scale as the diffusion 
estimate $\mathbf{x}_{0|t}$. This justifies the scale-aligned 
range-null composition in~\eqref{eq:range_null_recon}.

\subsection*{C. Discrete-Wavelet-Transform-Based Range-Null Operator}

The discrete wavelet transform low-pass (LL-subband) projection $\mathcal{A}$ satisfies
the Parseval-type energy identity
$\|\mathcal{A}\mathbf{x}\|_2^2 + \|(\mathbf{I}-\mathcal{A}^\dagger
\mathcal{A})\mathbf{x}\|_2^2 = \|\mathbf{x}\|_2^2$.
Under the physical premise that tissue structure is predominantly
low-frequency and speckle is predominantly high-frequency, we get:
\begin{equation}
  \mathcal{A}\mathbf{x} \approx \mathcal{A}\mathbf{y}, \qquad
  \mathcal{A}\mathbf{n} \approx \mathbf{0}.
\end{equation}
This motivates the range-null reconstruction in
\eqref{eq:range_null_recon}: the range-space component
$\mathcal{A}^\dagger\mathcal{A}\mathbf{y}_\mathrm{lin}$ anchors
the reconstruction to the observation, while the null-space
component $(\mathbf{I} - \mathcal{A}^\dagger\mathcal{A})\mathbf{x}_{0|t}$
is filled in by the diffusion prior.

\subsection*{D. Stability under Acoustic Distortion and Sound-Speed Inhomogeneity}

In practice, the received envelope deviates from the ideal
multiplicative model \eqref{eq:app_mult} due to:
frequency-dependent attenuation $a(z)$, which is partially corrected by
time-gain compensation; phase aberration loss
$\alpha_\mathrm{ab}$; sound-speed
inhomogeneity $\alpha_c$; and beamforming imperfections
$\alpha_\mathrm{bf}$. Because each factor is bounded and strictly positive, their product
\begin{equation}
  u_\mathrm{sys} = a(z)\,\alpha_\mathrm{ab}\,\alpha_c\,
                   \alpha_\mathrm{bf} > 0
\end{equation}
is a bounded positive multiplicative perturbation.
After log compression, this introduces an additive offset:
\begin{equation}
  \boldsymbol{\eta}_\mathrm{sys}
  = \frac{20}{D}\log_{10}(u_\mathrm{sys}).
\end{equation}
Thus, the full log-domain observation becomes
$\mathbf{y} = \mathbf{x} + \mathbf{n} +
\boldsymbol{\eta}_\mathrm{sys}$,
consistent with \eqref{eq:add_model} in the main text.
Crucially, $u_\mathrm{sys}>0$ means $\boldsymbol{\eta}_\mathrm{sys}$
is a deterministic offset and \emph{does not invalidate the additive
structure}: the log-domain additivity of \eqref{eq:app_exact}
is preserved up to the bounded term $\boldsymbol{\eta}_\mathrm{sys}$.

\subsection*{E. Upper Bounds on the Residual}

Let $B > 0$ denote the element-wise supremum of
$|\log u_\mathrm{sys}|$ in natural units.
We describe four nested upper bounds:

\textit{(E.1) Log-domain bound.} Since
$[\boldsymbol{\eta}_\mathrm{sys}]_i=\tfrac{20}{D}\log_{10}([u_\mathrm{sys}]_i)$
and $|\ln u_\mathrm{sys}|\leq B$ element-wise,
\begin{equation}
  \|\boldsymbol{\eta}_\mathrm{sys}\|_\infty
  \;\leq\; \frac{20\,B}{D\,\ln 10}.
  \label{eq:bound_log}
\end{equation}
For example, with $D = 60\,$dB and $\pm 3\,$dB peak distortion
($B = \tfrac{3}{20}\ln 10$), the right-hand side equals
$\tfrac{3}{60} = 0.05$ (5\% of the normalized full scale).

\textit{(E.2) Proxy-domain bound.}
Since $g^{-1}(u)=10^{D(u-1)/20}$, the ratio of the
inverse-compressed observation to the inverse-compressed
noiseless-plus-speckle signal satisfies the algebraic identity
\begin{equation}
  \frac{[g^{-1}(\mathbf{y})]_i}
       {[g^{-1}(\mathbf{x}+\mathbf{n})]_i}
  \;=\; 10^{\,D[\boldsymbol{\eta}_\mathrm{sys}]_i/20}
  \;=\; [u_\mathrm{sys}]_i,
  \label{eq:bound_proxy_step}
\end{equation}
because $[\boldsymbol{\eta}_\mathrm{sys}]_i =
\tfrac{20}{D}\log_{10}([u_\mathrm{sys}]_i)$ implies
$D[\boldsymbol{\eta}_\mathrm{sys}]_i/20 =
\log_{10}([u_\mathrm{sys}]_i)$ exactly.
Since $|\ln([u_\mathrm{sys}]_i)|\leq B$ implies
$\mathrm{e}^{-B}\leq[u_\mathrm{sys}]_i\leq \mathrm{e}^{B}$ and $\mathrm{e}^{B}-1 > 1-\mathrm{e}^{-B}$
for all $B>0$, the pointwise multiplicative distortion satisfies
\begin{equation}
  \left|\frac{[g^{-1}(\mathbf{y})]_i}
             {[g^{-1}(\mathbf{x}+\mathbf{n})]_i}
  - 1\right|
  \;=\; \bigl|[u_\mathrm{sys}]_i - 1\bigr|
  \;\leq\; \mathrm{e}^{B} - 1.
  \label{eq:bound_proxy}
\end{equation}
For example, with $D=60\,$dB and $\pm3\,$dB peak distortion
($B=3\ln 10/20$), this gives
$\mathrm{e}^{B}-1=10^{3/20}-1\approx 0.41$
(i.e., at most $41\%$ pointwise amplitude distortion).

\textit{(E.3) Range-space leakage bound.}
Denoting the range-null error by
$\boldsymbol{\delta} = \mathcal{A}\mathbf{n}
+ \mathcal{A}\boldsymbol{\eta}_\mathrm{sys}$:
\begin{equation}
  \|\boldsymbol{\delta}\|_2
  \;\leq\;
  \|\mathcal{A}\mathbf{n}\|_2
  + \|\mathcal{A}\|_{\mathrm{op}}
    \,\|\boldsymbol{\eta}_\mathrm{sys}\|_\infty\,
    \sqrt{HW}.
\end{equation}
The first term is the speckle leakage into the range space,
which is small by the frequency-separation assumption; while
the second term is the system-distortion bias.

\textit{(E.4) End-to-end UGNS update bound.}
Combining \eqref{eq:bound_proxy} and the $N$-ensemble variance
reduction (the standard deviation of $\hat{\boldsymbol{\mu}}$
scales as $\sigma/\sqrt{N}=\sqrt{\sigma^2/N}$), the deviation
of the UGNS output from the ideal noiseless estimate satisfies
\begin{equation}
  \bigl\|\hat{\mathbf{x}}_{0|t} - \hat{\mathbf{x}}_{0|t}^{\mathrm{ideal}}
  \bigr\|_2
  \;\leq\;
  \left(\sqrt{\frac{0.273}{N_\mathrm{eff}}}
        + \mathrm{e}^{B} - 1\right)
  \|\mathbf{s}_\mathrm{lin}\|_\infty,
  \label{eq:bound_e2e}
\end{equation}
where $N_\mathrm{eff} = N - 1$ for the unbiased variance
estimator and $\mathbf{s}_\mathrm{lin} = g^{-1}(\mathbf{x})$
is the noiseless positive-envelope proxy.
Bound \eqref{eq:bound_e2e} makes the modeling error explicit
and shows that it vanishes as $N \to \infty$ and $B \to 0$.


\begin{thebibliography}{00}
\bibitem{perrot2021think}
V. Perrot, M. Polichetti, F. Varray, and D. Garcia,
``So you think you can DAS? A viewpoint on delay-and-sum beamforming,''
\emph{Ultrasonics}, vol. 111, Mar. 2021, Art. no. 106309,
doi: 10.1016/j.ultras.2020.106309.

\bibitem{goodman1976some}
J. W. Goodman,
``Some fundamental properties of speckle,''
\emph{J. Opt. Soc. Amer.}, vol. 66, no. 11, pp. 1145--1150,
Nov. 1976, doi: 10.1364/JOSA.66.001145.

\bibitem{huang1979fast}
T. S. Huang, G. J. Yang, and G. Y. Tang,
``A fast two-dimensional median filtering algorithm,''
\emph{IEEE Trans. Acoust., Speech, Signal Process.}, vol. ASSP-27,
no. 1, pp. 13--18, Feb. 1979, doi: 10.1109/TASSP.1979.1163188.

\bibitem{yu2002srad}
Y. Yu and S. T. Acton,
``Speckle reducing anisotropic diffusion,''
\emph{IEEE Trans. Image Process.}, vol. 11, no. 11,
pp. 1260--1270, Nov. 2002, doi: 10.1109/TIP.2002.804276.

\bibitem{hyun2019beamforming}
D. Hyun, L. L. Brickson, K. T. Looby, and J. J. Dahl,
``Beamforming and speckle reduction using neural networks,''
\emph{IEEE Trans. Ultrason., Ferroelectr., Freq. Control},
vol. 66, no. 5, pp. 898--910, May 2019,
doi: 10.1109/TUFFC.2019.2903795.

\bibitem{luijten2020adaptive}
B. Luijten et al.,
``Adaptive ultrasound beamforming using deep learning,''
\emph{IEEE Trans. Med. Imag.}, vol. 39, no. 12,
pp. 3967--3978, Dec. 2020, doi: 10.1109/TMI.2020.3008537.

\bibitem{asgariandehkordi2023deep}
H. Asgariandehkordi, S. Goudarzi, A. Basarab, and H. Rivaz,
``Deep ultrasound denoising using diffusion probabilistic models,''
in \emph{Proc. IEEE Int. Ultrason. Symp. (IUS)},
Montreal, QC, Canada, Sep. 2023, pp. 1--4.

\bibitem{zhang2023reconstruction}
Y. Zhang, C. Huneau, J. Idier, and D. Mateus,
``Ultrasound image reconstruction with denoising diffusion restoration models,''
in \emph{Proc. 3rd MICCAI Workshop Deep Gener. Models (DGM4MICCAI)},
Vancouver, BC, Canada, 2023, pp. 193--203,
doi: 10.1007/978-3-031-53767-7\_19.

\bibitem{zhang2024variance}
Y. Zhang, C. Huneau, J. Idier, and D. Mateus,
``Ultrasound imaging based on the variance of a diffusion restoration model,''
in \emph{Proc. 32nd Eur. Signal Process. Conf. (EUSIPCO)},
Lyon, France, 2024, pp. 760--764.

\bibitem{jin2024lightweight}
Z. Xing et al.,
``A lightweight model for indoor object detection in unstructured scenes based on joint attention and prior knowledge in the context of home rehabilitation,''
\emph{J. King Saud Univ.-Comput. Inf. Sci.}, 2026,
doi: 10.1007/s44443-026-00605-w.

\bibitem{jin20253d}
Z. Xing, Z. Meng, G. Zheng, L. Yang, X. Guo, L. Tan, and Y. Jiang,
``Human-computer interactive rehabilitation: A 3D graph deep learning method for non-contact gesture recognition in post-epidemic and aging societies,''
\emph{Measurement}, 2025, Art. no. 118794,
doi: 10.1016/j.measurement.2025.118794.

\bibitem{kawar2022ddrm}
B. Kawar, M. Elad, S. Ermon, and J. Song,
``Denoising diffusion restoration models,''
in \emph{Proc. Adv. Neural Inf. Process. Syst. (NeurIPS)}, vol. 35,
2022, pp. 23593--23606.

\bibitem{lee1980digital}
J.-S. Lee,
``Digital image enhancement and noise filtering by use of local statistics,''
\emph{IEEE Trans. Pattern Anal. Mach. Intell.}, vol. PAMI-2,
no. 2, pp. 165--168, Mar. 1980, doi: 10.1109/TPAMI.1980.4766994.

\bibitem{loizou2005comparative}
C. P. Loizou, C. S. Pattichis, C. I. Christodoulou, R. S. H. Istepanian,
M. Pantziaris, and A. Nicolaides,
``Comparative evaluation of despeckle filtering in ultrasound imaging of the carotid artery,''
\emph{IEEE Trans. Ultrason., Ferroelectr., Freq. Control},
vol. 52, no. 10, pp. 1653--1669, Oct. 2005,
doi: 10.1109/TUFFC.2005.1561621.

\bibitem{buades2011nonlocal}
A. Buades, B. Coll, and J.-M. Morel,
``Non-local means denoising,''
\emph{Image Process. On Line}, vol. 1, pp. 208--212, 2011,
doi: 10.5201/ipol.2011.bcm\_nlm.

\bibitem{dabov2007image}
K. Dabov, A. Foi, V. Katkovnik, and K. Egiazarian,
``Image denoising by sparse 3-D transform-domain collaborative filtering,''
\emph{IEEE Trans. Image Process.}, vol. 16, no. 8,
pp. 2080--2095, Aug. 2007, doi: 10.1109/TIP.2007.901238.

\bibitem{asl2010eigenspace}
B. M. Asl and A. Mahloojifar,
``Eigenspace-based minimum variance beamforming applied to medical ultrasound imaging,''
\emph{IEEE Trans. Ultrason., Ferroelectr., Freq. Control},
vol. 57, no. 11, pp. 2381--2390, Nov. 2010,
doi: 10.1109/TUFFC.2010.1706.

\bibitem{bell2020cubdl}
M. A. L. Bell et al.,
``Challenge on ultrasound beamforming with deep learning (CUBDL),''
in \emph{Proc. IEEE Int. Ultrason. Symp. (IUS)},
Las Vegas, NV, USA, 2020, pp. 1--5.

\bibitem{goudarzi2020MobileNetV2}
S. Goudarzi, A. Asif, and H. Rivaz,
``Ultrasound beamforming using MobileNetV2,''
in \emph{Proc. IEEE Int. Ultrason. Symp. (IUS)},
Las Vegas, NV, USA, 2020, pp. 1--4.

\bibitem{lehtinen2018noise2noise}
J. Lehtinen et al.,
``Noise2Noise: Learning image restoration without clean data,''
in \emph{Proc. Int. Conf. Mach. Learn. (ICML)},
Stockholm, Sweden, 2018, pp. 2965--2974.

\bibitem{krull2019noise2void}
A. Krull, T.-O. Buchholz, and F. Jug,
``Noise2Void---Learning denoising from single noisy images,''
in \emph{Proc. IEEE/CVF Conf. Comput. Vis. Pattern Recognit. (CVPR)},
Long Beach, CA, USA, 2019, pp. 2124--2132.

\bibitem{cakmak2025lightweight}
M. Cakmak,
``A new lightweight hybrid model for pistachio classification using Transformers and EfficientNet,''
\emph{IEEE Access}, vol. 13, pp. 85857--85872, May 2025,
doi: 10.1109/ACCESS.2025.3567774.

\bibitem{ozel2025classification}
M. B. Ozel, S. B. Ay Kartbak, and M. Cakmak,
``Classification performance of deep learning models for the assessment of vertical dimension on lateral cephalometric radiographs,''
\emph{Diagnostics}, vol. 15, no. 17, Sep. 2025, Art. no. 2240,
doi: 10.3390/diagnostics15172240.

\bibitem{paral2025adaptive}
P. Paral, S. Ghosh, S. K. Pal, and A. Chatterjee,
``Adaptive non-homogeneous granulation-aided density-based deep feature clustering for far infrared sign language images,''
\emph{IEEE Trans. Emerg. Topics Comput. Intell.}, vol. 9, no. 2,
pp. 1269--1280, Apr. 2025, doi: 10.1109/TETCI.2024.3510292.

\bibitem{ghosh2023histogram}
S. Ghosh, P. Paral, A. Chatterjee, and S. Munshi,
``Histogram refined local ternary pattern-based bilateral LPP for vision sensor-based robot navigation guidance under challenging environments,''
\emph{IEEE Sensors Lett.}, vol. 7, no. 6, pp. 1--4, Jun. 2023,
doi: 10.1109/LSENS.2023.3272832.

\bibitem{guha2023sddpm}
S. Guha and S. T. Acton,
``SDDPM: Speckle denoising diffusion probabilistic models,''
2023, \emph{arXiv:2311.10868}.

\bibitem{li2025speckle2self}
X. Li, N. Navab, and Z. Jiang,
``Speckle2Self: Self-supervised ultrasound speckle reduction without clean data,''
\emph{Med. Image Anal.}, vol. 106, Dec. 2025, Art. no. 103755,
doi: 10.1016/j.media.2025.103755.

\bibitem{wang2023ddnm}
Y. Wang, J. Yu, and J. Zhang,
``Zero-shot image restoration using denoising diffusion null-space model,''
in \emph{Proc. Int. Conf. Learn. Represent. (ICLR)},
Kigali, Rwanda, 2023.

\bibitem{burckhardt1978speckle}
C. B. Burckhardt,
``Speckle in ultrasound B-mode scans,''
\emph{IEEE Trans. Sonics Ultrason.}, vol. SU-25, no. 1,
pp. 1--6, Jan. 1978, doi: 10.1109/T-SU.1978.30978.

\bibitem{wagner1983statistics}
R. F. Wagner, S. W. Smith, J. M. Sandrik, and H. Lopez,
``Statistics of speckle in ultrasound B-scans,''
\emph{IEEE Trans. Sonics Ultrason.}, vol. SU-30, no. 3,
pp. 156--163, May 1983, doi: 10.1109/T-SU.1983.31404.

\bibitem{oppenheim1968nonlinear}
A. V. Oppenheim, R. W. Schafer, and T. G. Stockham,
``Nonlinear filtering of multiplied and convolved signals,''
\emph{Proc. IEEE}, vol. 56, no. 8, pp. 1264--1291,
Aug. 1968, doi: 10.1109/PROC.1968.6570.

\bibitem{achim2001novel}
A. Achim, A. Bezerianos, and P. Tsakalides,
``Novel Bayesian multiscale method for speckle removal in medical ultrasound images,''
\emph{IEEE Trans. Med. Imag.}, vol. 20, no. 8,
pp. 772--783, Aug. 2001, doi: 10.1109/42.946592.

\bibitem{jain1989fundamentals}
A. K. Jain,
\emph{Fundamentals of Digital Image Processing}.
Englewood Cliffs, NJ, USA: Prentice-Hall, 1989.

\bibitem{loupas1989adaptive}
T. Loupas, W. N. McDicken, and P. L. Allan,
``An adaptive weighted median filter for speckle suppression in medical ultrasonic images,''
\emph{IEEE Trans. Circuits Syst.}, vol. 36, no. 1,
pp. 129--135, Jan. 1989, doi: 10.1109/31.16565.

\bibitem{choi2021ilvr}
J. Choi, S. Kim, Y. Jeong, Y. Gwon, and S. Yoon,
``ILVR: Conditioning method for denoising diffusion probabilistic models,''
in \emph{Proc. IEEE/CVF Int. Conf. Comput. Vis. (ICCV)},
Montreal, QC, Canada, 2021, pp. 14347--14356.

\bibitem{song2020ddim}
J. Song, C. Meng, and S. Ermon,
``Denoising diffusion implicit models,''
in \emph{Proc. Int. Conf. Learn. Represent. (ICLR)}, 2021.

\bibitem{liebgott2016plane}
H. Liebgott, A. Rodriguez-Molares, F. Cervenansky, J. A. Jensen, and O. Bernard,
``Plane-wave imaging challenge in medical ultrasound,''
in \emph{Proc. IEEE Int. Ultrason. Symp. (IUS)},
Tours, France, 2016, pp. 1--4.

\bibitem{rodriguez2020gcnr}
A. Rodriguez-Molares et al.,
``The generalized contrast-to-noise ratio: A formal definition of contrast for quantitative ultrasound,''
\emph{IEEE Trans. Ultrason., Ferroelectr., Freq. Control},
vol. 67, no. 4, pp. 745--759, Apr. 2020,
doi: 10.1109/TUFFC.2019.2956855.

\bibitem{dhariwal2021diffusion}
P. Dhariwal and A. Nichol,
``Diffusion models beat GANs on image synthesis,''
in \emph{Proc. Adv. Neural Inf. Process. Syst. (NeurIPS)}, vol. 34,
2021, pp. 8780--8794.

\bibitem{stevens2025semantic}
T. S. W. Stevens, O. Nolan, and R. J. G. van Sloun,
``Semantic diffusion posterior sampling for cardiac ultrasound dehazing,''
2025, \emph{arXiv:2508.17326}.

\bibitem{zhang2023informative}
Y. Zhang, C. Huneau, J. Idier, and D. Mateus,
``Diffusion reconstruction of ultrasound images with informative uncertainty,''
2023, \emph{arXiv:2310.20618}.

\bibitem{lu2022dpmsolver}
C. Lu, Y. Zhou, F. Bao, J. Chen, C. Li, and J. Zhu,
``DPM-Solver: A fast ODE solver for diffusion probabilistic model sampling in around 10 steps,''
in \emph{Proc. Adv. Neural Inf. Process. Syst. (NeurIPS)}, vol. 35,
2022, pp. 5775--5787.

\bibitem{li2023qdiffusion}
X. Li, Y. Liu, L. Lian, H. Yang, Z. Dong, D. Kang, S. Zhang, and K. Keutzer,
``Q-Diffusion: Quantizing diffusion models,''
in \emph{Proc. IEEE/CVF Int. Conf. Comput. Vis. (ICCV)},
Paris, France, 2023, pp. 17535--17545.
\end{thebibliography}
\end{document}